\documentclass{article} 
\usepackage{iclr2027_conference,times}

\usepackage{amsmath,amsfonts,bm}

\def\eqref#1{equation~\ref{#1}}

\def\1{\bm{1}}

\DeclareMathAlphabet{\mathsfit}{\encodingdefault}{\sfdefault}{m}{sl}
\SetMathAlphabet{\mathsfit}{bold}{\encodingdefault}{\sfdefault}{bx}{n}

\usepackage{hyperref}
\usepackage{url}

\usepackage{enumitem}
\usepackage{graphicx}
\usepackage{diagbox}
\usepackage{makecell}
\usepackage{subcaption}
\usepackage{array}
\usepackage{booktabs}
\usepackage{makecell}

\title{Beyond Solo and Consistency: Vindicating Multi-Agent Debate via Conditional Progressive Pruning}

\author{%
\normalfont
\textbf{Ruosong Ye}$^{1}$, \textbf{Caiqi Zhang}$^{2}$, \textbf{Jiahao Li}$^{3}$, \textbf{Haijun Wu}$^{1}$, \textbf{Xiaolong Luo}$^{4}$, \textbf{Huiyuan Chen}$^{5}$, \\
\textbf{Yu Wang}$^{6}$, \textbf{Ying Chen}$^{7}$, \textbf{Zhenting Wang}$^{1}$, \textbf{Kai Mei}$^{1}$, \textbf{Yang Zhou}$^{1}$, \textbf{Dimitris N. Metaxas}$^{1,*}$ \\[4pt]
$^{1}$Rutgers University, New Brunswick \quad
$^{2}$University of Cambridge \quad
$^{3}$Tsinghua University \\
$^{4}$Harvard University \quad
$^{5}$Case Western Reserve University \quad
$^{6}$University of California San Diego \\
$^{7}$Carnegie Mellon University \\[2pt]
$^{*}$Corresponding author \\
\texttt{ruosong.ye@rutgers.edu, cz391@cam.ac.uk, dnm@cs.rutgers.edu}
}

\iclrfinalcopy 
\begin{document}

\maketitle

\begin{abstract}
Large Language Model (LLM) based Multi-Agent Debate (MAD) is one of the most effective test time scaling techniques. Through multi-round communication, agents complement each other in knowledge and reasoning and solve tasks that no single member can solve. However, existing MAD frameworks fail to beat strong Single Agent and Consistency-based baselines under the same strict cost limit, which shakes the foundation of the MAD field. We propose Conditional Progressive Pruning (CPP), a lightweight pruning framework that fully exploits multi-round MAD. CPP outperforms all existing MAD frameworks on multiple dominated benchmarks. It is also the first to fully outperform consistency methods. Our code, detailed agent interaction records will be released soon.
\end{abstract}

\section{Introduction}
In today's AI field, the trend from individual intelligence to collective intelligence is becoming clearer \citep{du2023improving}, making Multi-Agent Collaboration one of the most popular research directions. Previous work has shown that collaboration is effective if and only if external information exists. Therefore we focus on the setting of heterogeneous LLM teams. The first type of collaboration work assigns different roles and duties to multiple agents \citep{liu2023dynamic}. These works essentially study task planning and distribution. Each agent still solves a single-step subtask alone. The second type of collaboration work can be called Multi-Agent Debate (MAD). In MAD, multiple agents are asked to solve the same problem together through multi-round communication \citep{li2026ofa}. Different LLMs complement each other in knowledge and reasoning. Thanks to this, a correct solution can emerge during communication even if no team member has a fully correct view at the start \citep{choi2026debate}. This means MAD cannot simply be seen as a process where strong agents persuade weak ones. In fact, the advantage of multi-round communication is not stable \citep{huang2024large}. Under the same computational cost, existing MAD frameworks can hardly fully outperform strong consistency-based baselines. Consistency methods first let different LLMs solve the problem independently. They then use majority voting to get the answer. As the number of agents increases, these methods show a stable trend of performance growth \citep{li2024more}. However, consistency methods are only independent repeated sampling. They are a communication-free, one-dimensional form of cooperation and their performance will eventually converge \citep{zhang2023exploring}. In contrast, agents in MAD can essentially improve reasoning through communication. So we believe the potential of MAD has not been fully exploited yet. That is why the value of the MAD field is still debated. Existing MAD frameworks generally have the following drawbacks: 
\begin{enumerate}[label=\arabic*), labelindent=1.5em, leftmargin=*, itemsep=2pt, topsep=4pt]
    \item Communication cost is counted only by the number of tokens, without normalizing the cost across different LLMs, so the cost is optimistically estimated.
    \item Did not make full use of the intermediate results of multi-round discussion for pruning, and they do not properly model the behavior of different agents to enable personalized information passing.
    \item They do not design reinforcement learning algorithms specifically for the debate setting, which leaves room for optimization.
\end{enumerate}
To address the above challenges, we propose Conditional Progressive Pruning (CPP), a lightweight pruning framework. Previous MAD methods make decisions only based on the interactions between LLM profiles \citep{chai2024expert}. Unlike them, CPP is the first to model the conditional probability between earlier and later actions in an autoregressive way. It also performs progressive pruning in the inference pipeline directly based on the results between rounds of MAD. This allows CPP to make full use of all the information produced in MAD for decision making, and thus shows the advantage of communication-based collaboration. The only work similar to ours is AnyMAC \citep{wang2025anymac}, which also successfully uses intermediate results to improve MAD performance. However, it does not model the relations between actions within and across rounds. Overall, our contributions can be summarized in the following three points:
\begin{itemize}[label=\textbullet, labelindent=1.5em, leftmargin=*, itemsep=2pt, topsep=4pt]
    \item To the best of our knowledge, we are the first to propose a framework that autoregressively prunes MAD based on intermediate results. It is also the first to fully outperform consistency methods and single agents under strictly the same computational cost. This ends the debate on whether MAD is valuable and vindicates the significance of the MAD field.
    \item We design a reinforcement learning algorithm for multi-round MAD. The proposed tailored reward and Agent Behavior Embedding can generally improve the convergence performance of MAD.
    \item Extensive experiments on MATH-Hard, GPQA and MMLU-Pro show that our CPP becomes the SoTA MAD framework, with consistent advantages in robustness, efficiency and scalability. We will also release the source code and detailed agent debate logs.
\end{itemize}

\section{Related Work}
\label{related}
\subsection{Heuristic Multi-Agent Debate}
Homogeneous and Heterogeneous Consistency Methods improve performance as the number of agents scales, through independent sampling and majority voting \citep{wang2022self}. They are a fully communication-free form of cooperation. Methods such as IoE \citep{li2024confidence}, PHP \citep{zheng2023progressive} and Reflexion \citep{shinn2023reflexion} let an agent discuss with itself over multiple rounds to reach a solution, but they are very unstable without external feedback \citep{choi2026debate}.  \citep{du2023improving} first proposes a static MAD framework in which agents are fully connected. Methods such as LLM-Intervention and $S^{2}$MAD \citep{zeng2025s2} then try to prune dynamically based on the quality and diversity of debate responses. Some works also let LLMs select information by themselves in a zero-shot way \citep{kim2024mdagents}, but this greatly increases the cost.

\subsection{Learning-based Multi-Agent Debate}
GPTSwarm \citep{zhuge2024language} is the first to introduce reinforcement learning for dynamic pruning in MAD. It formulates each communication edge as a single parameter, and works such as AgentPrune \citep{zhang2025cut} and AgentDropout \citep{wang2025agentdropout} are essentially the same. G-Designer \citep{zhang2024g} is the first to introduce an extra lightweight pruning model to optimize the flow, but its decisions rely only on the interaction between the profiles of two agents. MAS-GPT \citep{ye2025mas} is the first to generate MAD topologies in an autoregressive and generative way under supervised learning. ARG-Designer \citep{li2026assemble} later realizes this paradigm better through reinforcement learning. However, both of them rely only on agent embedding information. AnyMAC \citep{wang2025anymac} innovatively filters the results between rounds directly to achieve independent pruning. However, it does not consider the conditional probability between actions, and it does not explore reinforcement learning algorithms designed for debate. MasHost \citep{yang2025mashost} designs a reward for the debate setting, which gives rewards based on the change of the MAD state before and after an action. However, its numerical design does not fit the multi-round MAD setting. For agent embedding modeling, Symbolic-MoE \citep{chen2025symbolic} builds on text-introduction based profile initialization and adds statistical information. This is the simplest form of agent behavior modeling.


\section{CPP: Conditional Progressive Pruning}

\subsection{Inference Pipeline}

\begin{figure}[htbp]
    \centering
    \includegraphics[width=\linewidth]{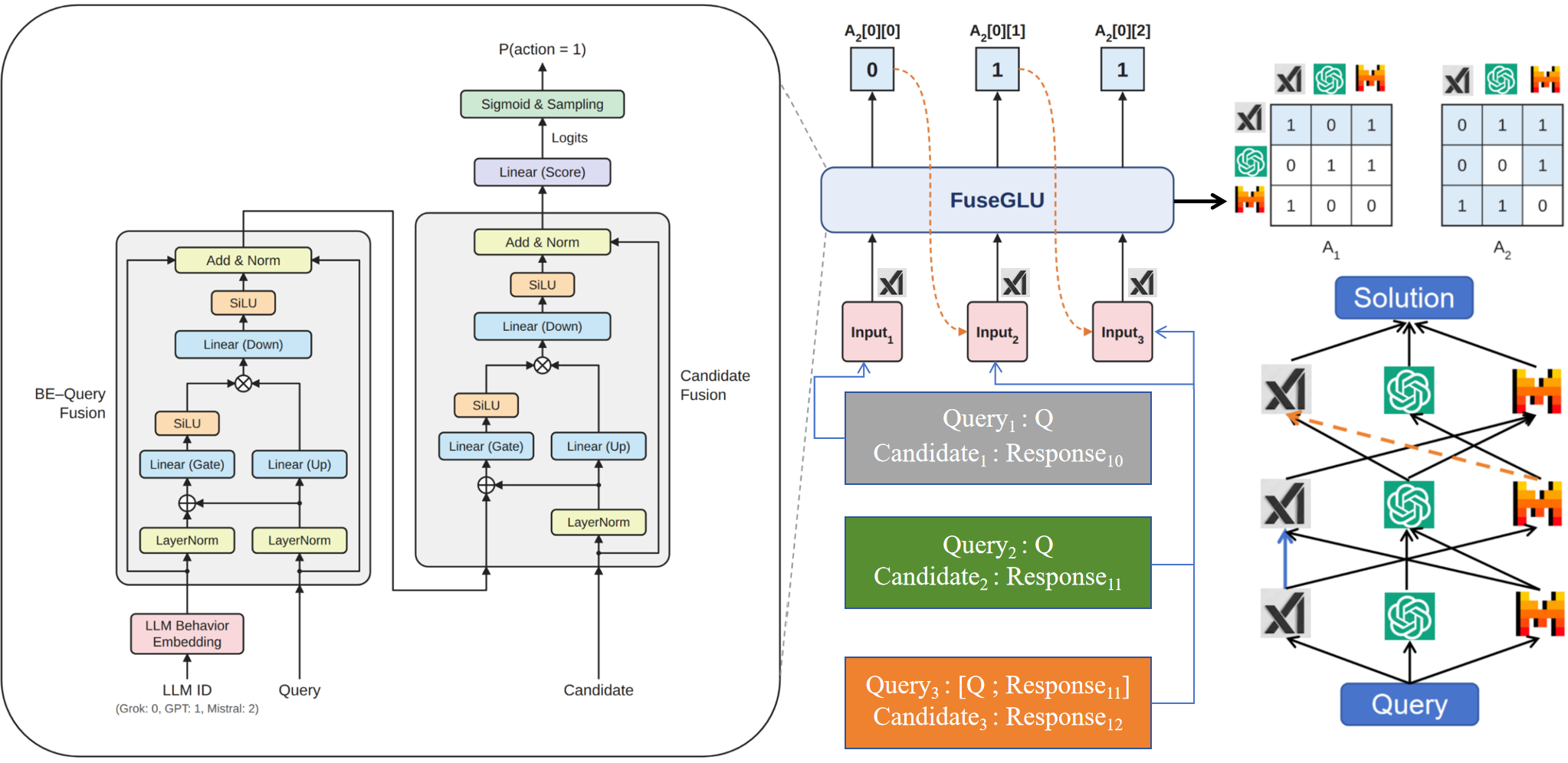}
    \caption{Overview of our CPP pipeline. FuseGLU takes the LLM behavior embedding, the query and the candidate response as input, and autoregressively decides each action by combining the available information with the candidate conditioned on previous actions.}
    \label{fig:pipeline}
\end{figure}


\subsubsection{Agent Behavior Modeling}
Previous works usually initialize agent embeddings with text-introduction profiles. In contrast, our behavior embedding is obtained by directly encoding the responses of each agent on specific questions from the validation set, and it is then trained together with CPP. This initialization directly models the concrete behavior of each agent, so it carries richer information than a general profile description.

\subsubsection{Conditional Pruning: FuseGLU}
Existing pruning models usually take the query as the main input and apply simple projections to it. In this case, the query is transformed with a low degree of nonlinearity, and the agent identity and the candidate response only interact with it in a shallow way. To address this, we propose FuseGLU, a variant of the Gated Linear Unit (GLU). Instead of transforming the query alone, FuseGLU fuses the agent behavior embedding and the candidate response into the gate and up projections \citep{yang2025qwen3}. This lets the extra information directly control how the query is transformed and increases the nonlinearity of the model. FuseGLU consists of two fusion blocks with the same form. For a main input $\mathbf{x}$ and a fused input $\mathbf{y}$, both after layer normalization, a fusion block is defined as
\begin{equation}
\mathrm{Fuse}(\mathbf{x},\mathbf{y})
=\mathrm{Norm}\Big(\mathbf{x}+\mathrm{SiLU}\big(W_{d}\big[\mathrm{SiLU}\big(W_{g}(\mathbf{x}+\mathbf{y})\big)\odot W_{u}\,\mathbf{y}\big]\big)\Big),
\label{eq:fuse}
\end{equation}
where $W_{g}$, $W_{u}$ and $W_{d}$ are the gate, up and down projections, and $\odot$ is element-wise multiplication. The first block fuses the behavior embedding $\mathbf{e}_i$ of the receiving agent $a_i$ with the query, and the second block fuses the result with the candidate response. The output is mapped to a score and gives the probability of keeping the edge from $a_j$ to $a_i$ in round $t$,
\begin{equation}
p_\theta\big(a^{t}_{i,j}=1\big)
=\sigma\Big(\mathbf{w}_{s}^{\top}\,\mathrm{Fuse}\big(\mathrm{Fuse}(\mathbf{e}_i,\mathbf{q}^{t}_{i,j}),\,\mathbf{r}^{t-1}_{j}\big)\Big),
\label{eq:prob}
\end{equation}
where $\mathbf{r}^{t-1}_{j}$ is the response of $a_j$ in the previous round and $\mathbf{q}^{t}_{i,j}$ is the current query.

The key of CPP is that the query is not fixed. Each agent first decides whether to keep its own previous response, and then decides on the other agents in a fixed order. Every response that has already been selected is concatenated to the original question, and the result becomes the query $\mathbf{q}^{t}_{i,j}$ for the next decision. In this way, each action is conditioned on the earlier actions of the same agent in the same round. Since the responses of round $t$ also depend on the edges chosen in round $t-1$, actions are linked across rounds as well. The whole pruning policy is therefore modeled autoregressively as
\begin{equation}
\pi_\theta\big(\mathcal{G}\big)
=\prod_{t=1}^{T}\prod_{i=1}^{N}\prod_{k=1}^{N}
p_\theta\big(a^{t}_{i,o_i(k)}\,\big|\,a^{t}_{i,o_i(<k)},\,\mathcal{H}^{t-1}\big),
\label{eq:ar}
\end{equation}
where $\mathcal{G}$ is the debate graph over all rounds, $o_i(k)$ is the $k$-th agent in the decision order of $a_i$, and $\mathcal{H}^{t-1}$ is the debate history up to round $t-1$. This makes full use of the intermediate results of MAD when making each pruning decision.

\subsection{Progressive Reinforcement Learning for MAD}

\subsubsection{Debate-Tailored Reward Design}
We design a reward tailored for multi-round debate, as shown in Table~\ref{tab:reward}. The sign pair indicates whether the answer of the current agent is correct before and after an action, and the action is 1 if the edge is kept and 0 if it is pruned. The scalar $a$ controls the scale of the reward. We use a larger $a$ in later rounds, since they are closer to the final result. When the answer is already correct ($++$), further communication is not necessary. So pruning the edge is rewarded with $a$, while keeping it is rewarded with $k\cdot a$, where $k\in[-1,1]$ is a coefficient that can make this reward either positive or negative. An action that turns a correct answer into a wrong one ($+-$) receives $-2a$, and an action that turns a wrong answer into a correct one ($-+$) receives $2a$. When the answer stays wrong ($--$), both actions receive $-a/(N-1)$, where $N$ is the number of agents in one round. This small penalty encourages agents to act in order to move from wrong to correct. Even if the answer stays wrong in all earlier actions of a round and only becomes correct at the last action, the accumulated reward of these actions is still positive.

\begin{table}[htbp]
\caption{State-Variation-Aware Multi-Round Debate Reward}
\label{tab:reward}
\begin{center}
\begin{tabular}{cccccc}
\hline
\diagbox{Action}{Sign} & \bf + + & \bf + - & \bf - + & \bf - - \\
\hline
1 & $k \cdot a$ & $-2a$ & $2a$ & $-a/(N-1)$ \\
0 & $a$ & -- & -- & $-a/(N-1)$ \\
\hline
\end{tabular}
\end{center}
\end{table}

\subsubsection{Inter-Round Reward Accumulation}
An action affects not only its own result but also the actions that follow it. Take the bold blue edge in the first round of Figure~\ref{fig:pipeline} as an example. This edge decides whether Grok keeps its own previous response. Its result is concatenated into the query of Grok's later decisions in the same round, so it affects Grok's choices on GPT and Mistral. It also changes Grok's response in this round, which becomes a candidate for the agents in the next round. The actions affected by this edge are marked with light blue backgrounds in $A_1$ and $A_2$. Therefore, the return of an action accumulates the rewards of all its subsequent related actions. We use two discount factors for this accumulation. $\beta$ discounts rewards across rounds, and $\gamma$ discounts rewards within a round. The subsequent actions of each action are listed in Appendix~\ref{AAA}. We then optimize the pruning policy with on-policy policy gradient,
\begin{equation}
\nabla_\theta J(\theta)
=\mathbb{E}_{\mathcal{G}\sim\pi_\theta}\Big[\sum_{t=1}^{T}\sum_{i=1}^{N}\sum_{k=1}^{N}
G^{t}_{i,o_i(k)}\,\nabla_\theta\log p_\theta\big(a^{t}_{i,o_i(k)}\,\big|\,a^{t}_{i,o_i(<k)},\,\mathcal{H}^{t-1}\big)\Big],
\label{eq:pg}
\end{equation}
where $G^{t}_{i,o_i(k)}$ is the accumulated return of action $a^{t}_{i,o_i(k)}$ discounted by $\beta$ and $\gamma$.

\section{Experiments}

\subsection{Experimental Setup}
We conduct experiments on benchmarks such as MATH-Hard \citep{hendrycks2021measuring}, MMLU-Pro \citep{wang2024mmlu} and GPQA-Diamond \citep{rein2023gpqa}, which test the reasoning ability, the factuality ability and the combination of both in MAD, respectively. For MATH-Hard, we sample 150/100/100 problems of the highest difficulty level (Level 5) from MATH as the training/validation/test sets. The corresponding sample sizes are 150/100/100 for MMLU-Pro and 98/50/100 for GPQA. All main experiments are done with Grok-4.1-non-reasoning, GPT-4.1-nano and Mistral-small-3.2. Due to the deprecation of Grok 4.1, we switch to Grok-4.3 for the robustness experiments, the ablation studies on the reward and the pipeline, and the efficiency experiments. Our pruning model uses the proposed FuseGLU architecture, and all experiments are trained with on-policy reinforcement learning. The main hyperparameters we tune include the learning rate, batch size, entropy coefficient and entropy decay rate. More details will be provided in our code, which will be open-sourced soon. More detailed information about all baselines can be found in Section~\ref{related}


\subsection{Main Results}

\subsubsection{Beyond Existing MADs}
Table~\ref{tab:main} compares CPP with existing MAD methods, where the cost is normalized with respect to Full-Debate. CPP achieves the highest accuracy on all three benchmarks, reaching 68.17\% on MATH-Hard, 68.67\% on GPQA-Diamond and 66.33\% on MMLU-Pro. At the same time, its cost stays below that of Full-Debate on every benchmark. The strongest baselines, ARG-Designer and AnyMAC, also use learned pruning, but CPP still outperforms both of them on all benchmarks, with the largest gain on MMLU-Pro. These results show that CPP outperforms all existing MAD frameworks. 

\begin{table}[htbp]
\caption{CPP Main Results: Comparison with MADs.}
\label{tab:main}
\begin{center}
\resizebox{\linewidth}{!}{%
\begin{tabular}{lcc|cc|cc}
\hline
\multicolumn{1}{c}{} & \multicolumn{2}{c}{\bf MATH-Hard} & \multicolumn{2}{c}{\bf GPQA-Diamond} & \multicolumn{2}{c}{\bf MMLU-Pro} \\
\cline{2-3} \cline{4-5} \cline{6-7}
\multicolumn{1}{c}{\bf Method} &\multicolumn{1}{c}{Acc (\%)} &\multicolumn{1}{c}{Cost} &\multicolumn{1}{c}{Acc (\%)} &\multicolumn{1}{c}{Cost} &\multicolumn{1}{c}{Acc (\%)} &\multicolumn{1}{c}{Cost}
\\ \hline
Grok              & 48.50 $\pm$ 1.38 & 11.17\%  & 57.33 $\pm$ 4.27 & 13.50\%  & 57.00 $\pm$ 2.90 & 12.05\%  \\
GPT               & 43.83 $\pm$ 4.79 & 7.56\%   & 38.67 $\pm$ 3.78 & 6.78\%   & 42.17 $\pm$ 3.43 & 6.49\%   \\
Mistral           & 46.17 $\pm$ 3.97 & 8.15\%   & 45.33 $\pm$ 3.44 & 4.46\%   & 57.33 $\pm$ 1.51 & 5.22\%   \\
Vote              & 55.00 $\pm$ 3.63 & 26.88\%  & 58.67 $\pm$ 3.01 & 24.74\%  & 60.67 $\pm$ 3.61 & 23.75\%  \\
\hline
IoE-Grok          & 56.83 $\pm$ 3.06 & 28.24\%  & 65.83 $\pm$ 2.32 & 32.79\%  & 58.83 $\pm$ 3.60 & 32.84\%  \\
IoE-GPT           & 48.50 $\pm$ 4.85 & 17.12\%  & 45.33 $\pm$ 5.47 & 15.48\%  & 47.67 $\pm$ 1.63 & 15.30\%  \\
IoE-Mistral       & 47.33 $\pm$ 3.44 & 20.83\%  & 45.83 $\pm$ 3.76 & 11.21\%  & 53.67 $\pm$ 2.66 & 11.84\%  \\
IoE-Vote          & 60.33 $\pm$ 3.72 & 66.19\%  & 62.50 $\pm$ 1.97 & 59.48\%  & 61.17 $\pm$ 3.31 & 59.98\%  \\
php-Grok          & 53.83 $\pm$ 1.94 & 27.68\%  & 61.00 $\pm$ 2.61 & 28.85\%  & 57.00 $\pm$ 2.00 & 26.24\%  \\
php-GPT           & 51.00 $\pm$ 3.46 & 17.83\%  & 42.50 $\pm$ 4.04 & 14.62\%  & 44.50 $\pm$ 3.08 & 14.17\%  \\
php-Mistral       & 50.83 $\pm$ 2.99 & 22.50\%  & 40.50 $\pm$ 1.87 & 9.24\%   & 57.83 $\pm$ 1.17 & 10.83\%  \\
php-Vote          & 59.33 $\pm$ 2.73 & 68.00\%  & 60.17 $\pm$ 4.02 & 52.70\%  & 60.50 $\pm$ 1.38 & 51.23\%  \\
Reflexion-Grok    & 54.17 $\pm$ 2.71 & 36.81\%  & 66.83 $\pm$ 2.23 & 53.31\%  & 62.17 $\pm$ 2.64 & 52.63\%  \\
Reflexion-GPT     & 47.17 $\pm$ 2.23 & 20.53\%  & 49.33 $\pm$ 3.27 & 30.26\%  & 47.83 $\pm$ 3.13 & 26.73\%  \\
Reflexion-Mistral & 51.17 $\pm$ 1.33 & 16.95\%  & 46.50 $\pm$ 3.62 & 20.35\%  & 56.83 $\pm$ 1.47 & 22.13\%  \\
Reflexion-Vote    & 60.83 $\pm$ 3.66 & 74.29\%  & 63.83 $\pm$ 3.19 & 103.92\% & 63.17 $\pm$ 2.40 & 101.48\% \\
\hline
Full-Debate       & 65.67 $\pm$ 3.27 & \textbf{100.00\%} & 67.50 $\pm$ 2.88 & \textbf{100.00\%} & 63.67 $\pm$ 1.86 & \textbf{100.00\%} \\
Random            & 64.50 $\pm$ 5.05 & 91.39\%  & 65.33 $\pm$ 4.08 & 91.90\%  & 64.17 $\pm$ 4.26 & 90.01\%  \\
LLM-Self          & 64.00 $\pm$ 2.10 & 119.32\% & 66.50 $\pm$ 4.72 & 138.00\% & 61.83 $\pm$ 4.36 & 141.02\% \\
Intervention      & 63.67 $\pm$ 2.66 & 91.97\%  & 61.17 $\pm$ 3.97 & 94.28\%  & 60.50 $\pm$ 2.88 & 90.89\%  \\
$S^2$-MAD         & 60.00 $\pm$ 2.53 & 96.72\%  & 62.17 $\pm$ 4.62 & 93.89\%  & 61.50 $\pm$ 2.07 & 92.30\%  \\
IoE-Debate        & 64.17 $\pm$ 3.87 & 92.01\%  & 68.17 $\pm$ 2.79 & 92.24\%  & 62.67 $\pm$ 1.51 & 93.97\%  \\
Debate-IoE        & 65.67 $\pm$ 2.58 & 94.24\%  & 67.00 $\pm$ 3.63 & 90.16\%  & 60.83 $\pm$ 4.36 & 93.37\%  \\
Zero              & 57.33 $\pm$ 2.73 & 84.93\%  & 58.67 $\pm$ 3.27 & 81.50\%  & 63.50 $\pm$ 2.88 & 79.37\%  \\
\hline
ARG-Designer      & 68.00 $\pm$ 4.38 & 99.58\%  & 67.67 $\pm$ 3.27 & 97.09\%  & 64.00 $\pm$ 4.38 & 86.90\%  \\
AnyMac            & 67.83 $\pm$ 2.32 & 95.94\%  & 65.67 $\pm$ 2.80 & 94.40\%  & 63.00 $\pm$ 2.53 & 90.27\%  \\
\textbf{CPP}      & \textbf{68.17 $\pm$ 2.14} & 98.61\%  & \textbf{68.67 $\pm$ 1.51} & 97.63\%  & \textbf{66.33 $\pm$ 3.33} & 99.43\%  \\
\hline
\end{tabular}%
}
\end{center}
\end{table}

Beyond the top-line comparison, the table reveals several consistent patterns. Majority voting (Vote) already provides a sizable boost over any single model (e.g., 55.00\% vs.\ at most 48.50\% on MATH-Hard), but the self-refinement family (IoE, php, Reflexion) achieves further gains only at a steep cost premium---Reflexion-Vote, for instance, exceeds Full-Debate's own cost on both MATH-Hard (74.29\%) and GPQA-Diamond (103.92\%) while still trailing CPP in accuracy by 4--8 points. Among the fixed-topology MAD baselines, cost varies substantially (85\%--141\% across benchmarks) without a corresponding accuracy advantage: LLM-Self incurs the highest cost on two benchmarks yet remains below CPP in accuracy, and $S^2$-MAD combines moderate cost with the lowest accuracy in its group, indicating that neither self-determined nor heuristic pruning reliably identifies efficient communication structures. Further detailed referenced ratio and edge statistics can be found in Appendix~\ref{abcde}

Comparing the two learned-pruning baselines with CPP further highlights the value of the proposed design. ARG-Designer is competitive on MATH-Hard (68.00\%) but its accuracy drops on MMLU-Pro (64.00\%) alongside a markedly lower cost (86.90\%), suggesting over-aggressive pruning on that benchmark; AnyMAC is more cost-stable (90--96\%) but consistently falls 0.3--3.3 points behind CPP in accuracy. CPP, in contrast, maintains both the highest accuracy and a cost consistently below Full-Debate (93.79--99.43\%) across all three benchmarks, indicating that its pruning decisions generalize better across task distributions rather than overfitting to a specific benchmark's structure.

\subsubsection{Beyond Extended Consistency}
Table~\ref{tab:cons} compares MAD methods with consistency methods under the same computational cost. Rows named after a single model use homogeneous consistency, and all other rows use heterogeneous consistency. Our debate has 3 agents and 3 rounds, which corresponds to 9 samples. However, communication brings extra cost, so giving consistency methods only 9 samples would not be a fair comparison. We therefore extend the number of samples of consistency methods beyond 9 to match the cost of MAD. Consistency-Best extends the samples using only Grok, the strongest model, while Consistency-Fuse extends the samples of all models in equal proportion.

The last three rows with the prefix E denote the Extension setting. In the main experiments, each round contains one Grok, one GPT and one Mistral agent. In the Extension setting, we change the arrangement so that the first round uses only Mistral, the second round uses only GPT and the third round uses only Grok. MAD methods still beat consistency methods under this new arrangement, which shows that MAD has a large room for further optimization.

Most importantly, CPP is the only method that outperforms all consistency methods on every benchmark. On MMLU-Pro, CPP in Table~\ref{tab:main} is the only MAD method that beats Consistency-Fuse. This remains true in the Extension setting, where E\_CPP is still the only method that outperforms all consistency methods on MMLU-Pro.

\begin{figure}[htbp]
\centering
\begin{subfigure}[b]{0.333\linewidth}
  \centering
  \includegraphics[width=\linewidth]{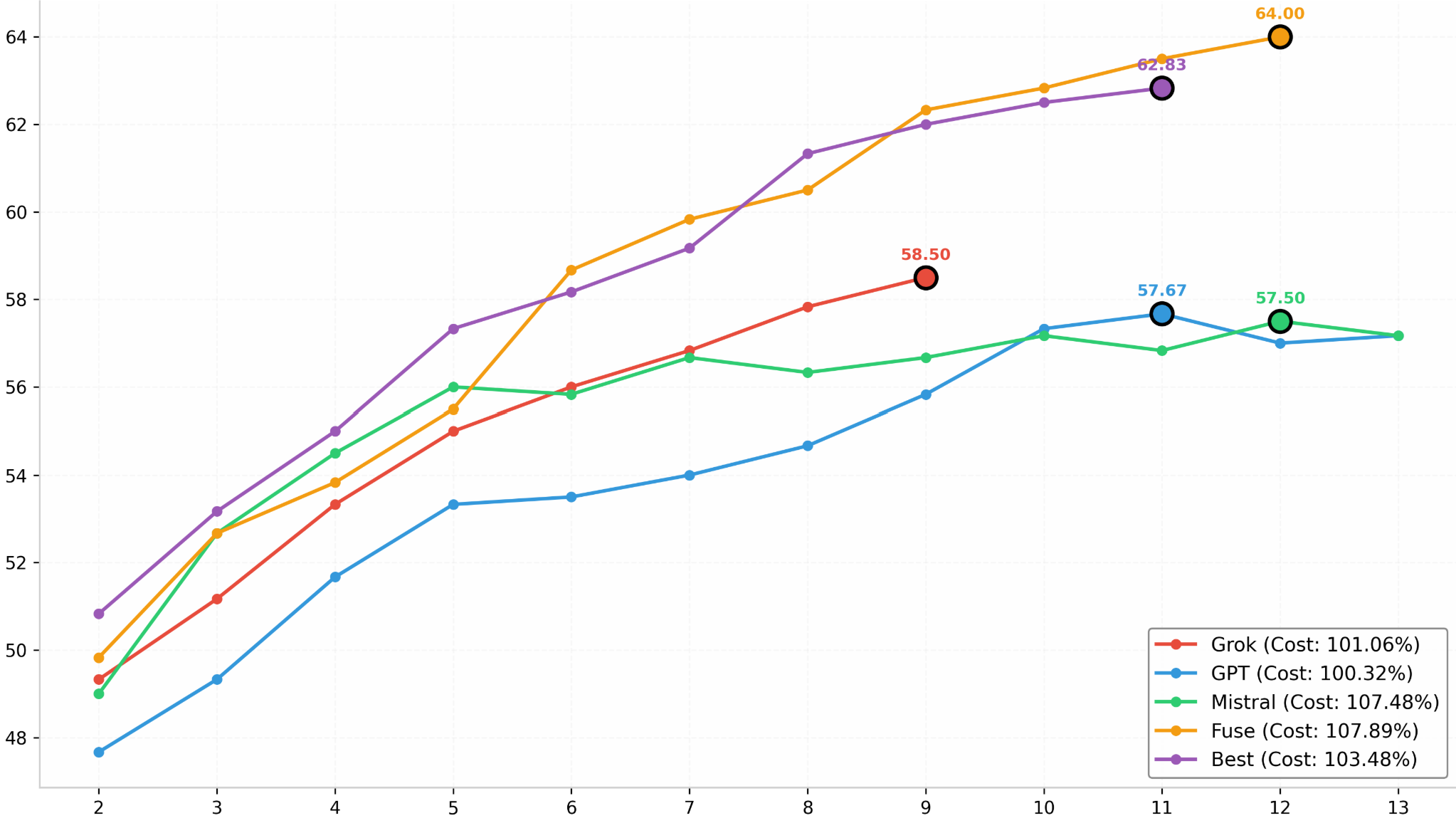}
  \caption{MATH}
  \label{fig:sub-a}
\end{subfigure}%
\begin{subfigure}[b]{0.333\linewidth}
  \centering
  \includegraphics[width=\linewidth]{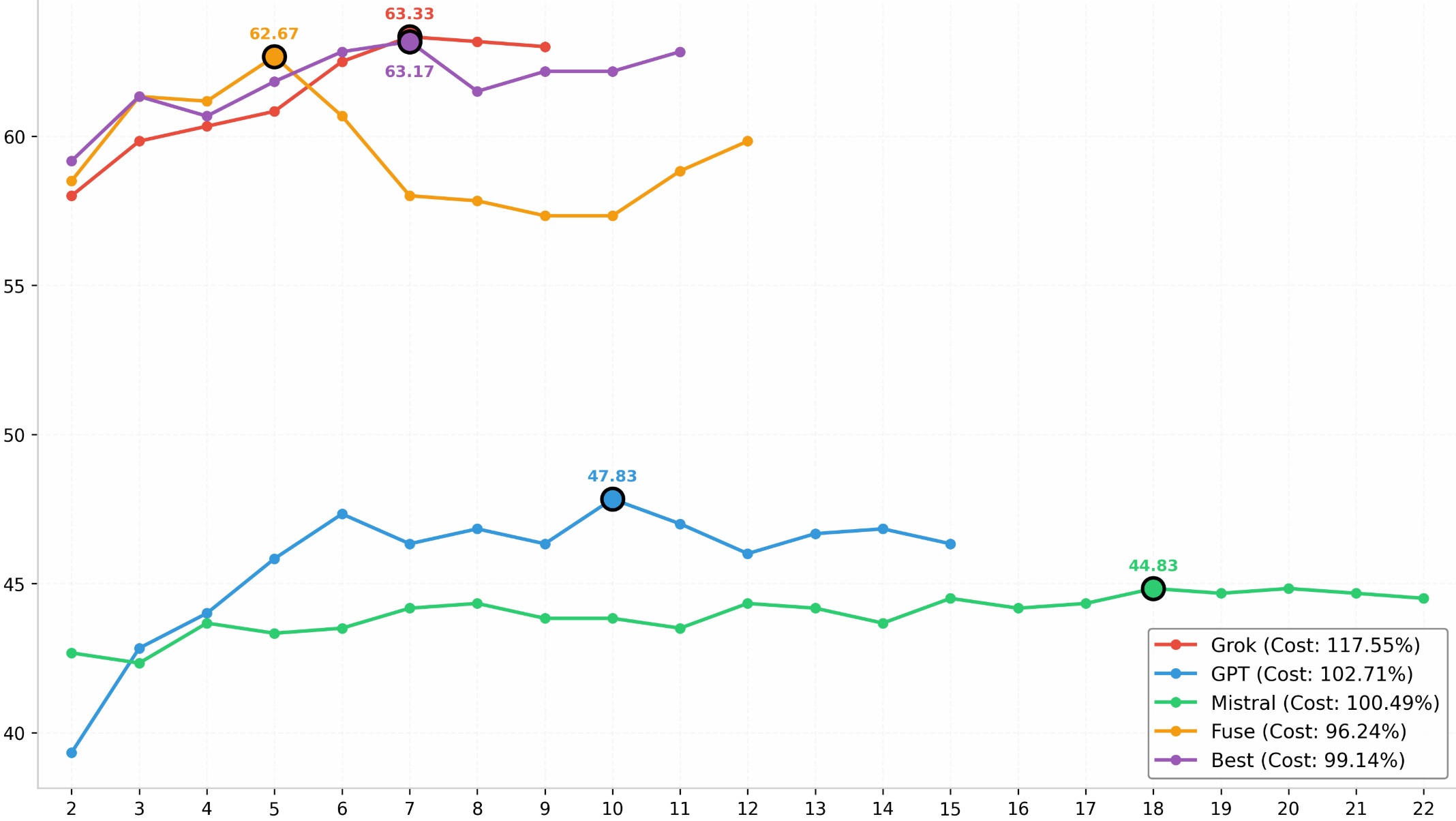}
  \caption{GPQA}
  \label{fig:sub-b}
\end{subfigure}%
\begin{subfigure}[b]{0.333\linewidth}
  \centering
  \includegraphics[width=\linewidth]{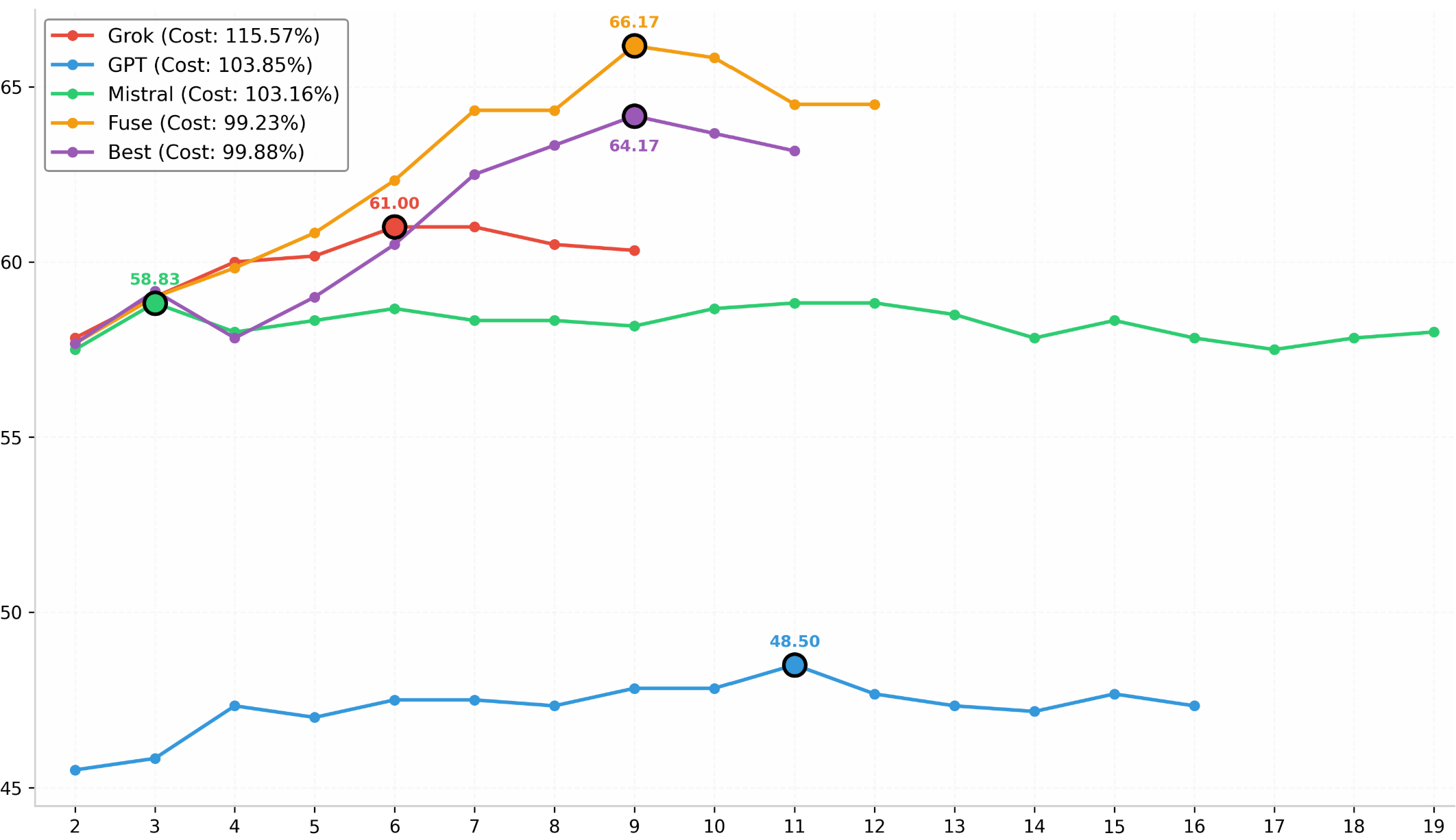}
  \caption{MMLU}
  \label{fig:sub-c}
\end{subfigure}
\caption{Detailed Consistency Results. Full results can be found in Appendix~\ref{dcons}}
\label{fig:cons}
\end{figure}

\begin{table}[htbp]
\caption{All Consistency Mode and MAD Struture Extension Results}
\label{tab:cons}
\begin{center}
\resizebox{\linewidth}{!}{%
\begin{tabular}{lcc|cc|cc}
\hline
\multicolumn{1}{c}{} & \multicolumn{2}{c}{\bf MATH-Hard} & \multicolumn{2}{c}{\bf GPQA-Diamond} & \multicolumn{2}{c}{\bf MMLU-Pro} \\
\cline{2-3} \cline{4-5} \cline{6-7}
\multicolumn{1}{c}{\bf Method} &\multicolumn{1}{c}{Acc (\%)} &\multicolumn{1}{c}{Cost} &\multicolumn{1}{c}{Acc (\%)} &\multicolumn{1}{c}{Cost} &\multicolumn{1}{c}{Acc (\%)} &\multicolumn{1}{c}{Cost}
\\ \hline
Consistency-Grok    & 58.50 $\pm$ 1.97 & 101.06\% & \underline{63.33 $\pm$ 1.63} & 91.43\% & 61.00 $\pm$ 1.55 & 77.05\%  \\
Consistency-GPT     & 57.67 $\pm$ 1.03 & 84.88\%  & 47.83 $\pm$ 1.60 & 68.47\% & 48.50 $\pm$ 1.05 & 71.40\%  \\
Consistency-Mistral & 57.50 $\pm$ 1.97 & 99.21\%  & 45.33 $\pm$ 3.44 & 4.46\%  & 58.83 $\pm$ 1.72 & 16.29\%  \\
\hline
Consistency-Fuse    & \underline{64.00 $\pm$ 2.45} & 107.89\% & 62.67 $\pm$ 2.16 & 56.88\% & \underline{66.17 $\pm$ 2.32} & 78.94\%  \\
Consistency-Best    & 62.83 $\pm$ 2.14 & 103.48\% & 63.17 $\pm$ 2.48 & 74.85\% & 64.17 $\pm$ 3.06 & 86.30\%  \\
\hline
E\_Full-Debate           & \textbf{69.50 $\pm$ 2.59} & 103.95\% & 66.33 $\pm$ 3.14 & 100.52\% & 66.00 $\pm$ 2.68 & 103.83\% \\
E\_ARG-Designer               & 68.17 $\pm$ 1.17 & 103.15\% & 64.00 $\pm$ 1.67 & 95.45\%  & 62.83 $\pm$ 2.14 & 85.21\%  \\
E\_CPP              & 67.17 $\pm$ 2.04 & 101.23\% & \textbf{68.50 $\pm$ 1.38} & 97.13\%  & \textbf{68.00 $\pm$ 2.28} & 99.12\%  \\
\hline
\end{tabular}%
}
\end{center}
\end{table}

\subsubsection{Beyond Scaled Chain of Thought}
\begin{table}[htbp]
\caption{\textbf{Accuracy}  (\%) and \textbf{Cost} of single models across different numbers of CoTs on MATH-Hard}
\label{tab:cots}
\begin{center}
\resizebox{\linewidth}{!}{%
\begin{tabular}{c|ccccccccc}
\hline
\diagbox{Model}{CoTs} & 0 & 6 & 9 & 15 & 21 & 27 & 36 & 48 & 60 \\
\hline
Grok    & \makecell{48.17$\pm$2.93\\5.65\%}  & \makecell{46.67$\pm$3.08\\17.44\%} & \makecell{46.50$\pm$2.59\\24.68\%} & \makecell{44.50$\pm$3.08\\40.78\%} & \makecell{44.33$\pm$2.66\\54.26\%} & \makecell{45.00$\pm$3.03\\67.35\%} & \makecell{44.67$\pm$1.86\\86.47\%} & \makecell{46.83$\pm$2.48\\110.28\%} & -- \\
\hline
GPT     & \makecell{28.17$\pm$2.14\\3.69\%}  & \makecell{\textbf{46.83$\pm$2.64}\\11.37\%} & \makecell{41.83$\pm$4.36\\14.92\%} & \makecell{42.83$\pm$3.54\\22.91\%} & \makecell{42.83$\pm$2.64\\29.53\%} & \makecell{40.50$\pm$3.15\\36.56\%} & \makecell{41.83$\pm$4.12\\46.74\%} & \makecell{42.67$\pm$3.01\\59.20\%}  & \makecell{41.33$\pm$4.72\\76.52\%} \\
\hline
Mistral & \makecell{18.17$\pm$4.58\\2.41\%}  & \makecell{44.17$\pm$3.13\\11.20\%} & \makecell{\textbf{51.00$\pm$3.41}\\16.58\%} & \makecell{47.00$\pm$5.66\\24.79\%} & \makecell{49.67$\pm$2.25\\33.53\%} & \makecell{48.50$\pm$3.27\\41.21\%} & \makecell{47.50$\pm$3.02\\51.00\%} & \makecell{49.67$\pm$1.63\\65.37\%}  & \makecell{47.67$\pm$2.94\\81.50\%} \\
\hline
\end{tabular}%
}
\end{center}
\end{table}
We also compare CPP with single agents that use more CoTs under a similar cost, as shown in Table~\ref{tab:cots}. Adding more CoTs does not bring steady gains to single agents, and the best result among all settings is only 51.00\% from Mistral with 9 CoTs. In contrast, CPP reaches 68.17\% on MATH-Hard with a cost of 98.61\%, which clearly outperforms single agents under the same computational cost.

\subsection{Ablation Study and Analysis}
\subsubsection{Adversarial Robustness}
Table~\ref{tab:adv} reports results after Grok is updated to Grok-4.3, under both the vanilla and the adversarial settings. In the vanilla setting, CPP achieves the best accuracy of 65.17\% with a lower cost than Full-Debate and ARG-Designer. In the adversarial setting, single agents and consistency methods drop sharply to around 20\% or lower, while debate methods remain much more robust. Since CPP makes decisions based on intermediate results, it can be directly trained in the adversarial setting. In contrast, methods such as ARG-Designer only rely on agent profiles and cannot perform adversarial training. After adversarial training, Adv-CPP keeps the same accuracy of 65.17\% in the vanilla setting, with a lower cost and a smaller variance. In the adversarial setting, its accuracy increases from 57.33\% to 58.00\%, which is the best among all methods.

\begin{table}[htbp]
\caption{Adversarial Robustness}
\label{tab:adv}
\begin{center}
\resizebox{\linewidth}{!}{%
\begin{tabular}{lcccc|cccc}
\hline
\multicolumn{1}{c}{} & \multicolumn{4}{c}{\bf Vanilla} & \multicolumn{4}{c}{\bf Adversarial} \\
\cline{2-5} \cline{6-9}
\multicolumn{1}{c}{\bf Method} &\multicolumn{1}{c}{Ref} &\multicolumn{1}{c}{Acc (\%)} &\multicolumn{1}{c}{Cost} &\multicolumn{1}{c}{Edge} &\multicolumn{1}{c}{Ref} &\multicolumn{1}{c}{Acc (\%)} &\multicolumn{1}{c}{Cost} &\multicolumn{1}{c}{Edge}
\\ \hline
Grok                & -- & 41.17 $\pm$ 3.60 & 13.24\% & -- & -- & 9.83 $\pm$ 2.48  & 9.78\%  & -- \\
GPT                 & -- & 43.83 $\pm$ 4.79 & 3.95\%  & -- & -- & 4.67 $\pm$ 2.34  & 1.01\%  & -- \\
Mistral             & -- & 46.17 $\pm$ 3.97 & 4.26\%  & -- & -- & 12.67 $\pm$ 2.07 & 1.07\%  & -- \\
Vote                & -- & 51.17 $\pm$ 3.31 & 21.45\% & -- & -- & 10.17 $\pm$ 1.72 & 11.86\% & -- \\
IoE-Vote            & -- & 52.83 $\pm$ 3.97 & 51.66\% & -- & -- & 47.17 $\pm$ 4.26 & 52.34\% & -- \\
\hline
Consistency-Mistral & -- & 59.67 $\pm$ 2.25 & 94.62\% & -- & -- & 20.83 $\pm$ 1.47 & 16.41\% & -- \\
Consistency-Fuse    & -- & 62.50 $\pm$ 0.84 & 94.94\% & -- & -- & 18.00 $\pm$ 1.26 & 54.67\% & -- \\
Consistency-Best    & -- & 62.17 $\pm$ 1.72 & 94.81\% & -- & -- & 18.17 $\pm$ 1.60 & 37.76\% & -- \\
\hline
Zero$^*$                & 0, 0, 0          & 52.50 $\pm$ 1.97 & 68.23\%  & 0.00  & 0.5, 0.5, 0.5    & \textit{\textbf{61.17 $\pm$ 2.93}} & 77.67\% & 9.00  \\
Full-Debate         & 1, 1, 1          & 62.17 $\pm$ 1.94 & \textbf{100.00\%} & 18.00 & 1, 1, 1          & 54.50 $\pm$ 1.64 & 78.28\% & 18.00 \\
Debate-IoE          & 0.67, 0.67, 0.67 & 61.00 $\pm$ 2.53 & 87.39\%  & 12.00 & 0.67, 0.67, 0.67 & 53.67 $\pm$ 2.16 & 62.67\% & 12.00 \\
IoE-Debate          & 0.67, 0.67, 0.67 & 59.67 $\pm$ 3.14 & 79.36\%  & 12.00 & 0.67, 0.67, 0.67 & 57.67 $\pm$ 2.58 & 72.64\% & 12.00 \\
\hline
ARG-Designer        & 0.99, 0.99, 0.99 & 64.67 $\pm$ 4.55 & 101.55\% & 17.79 & 0.99, 0.99, 0.99 & 56.83 $\pm$ 2.32 & 78.78\% & 17.80 \\
CPP                 & 0.91, 0.87, 0.86 & \underline{\textbf{65.17 $\pm$ 3.37}} & 93.79\%  & 15.83 & 0.9, 0.86, 0.86  & 57.33 $\pm$ 2.66 & 77.49\% & 15.73 \\
Adv-CPP             & 0.9, 0.86, 0.86  & \textbf{65.17 $\pm$ 1.47} & 91.50\%  & 15.70 & 0.89, 0.85, 0.85 & \underline{58.00 $\pm$ 2.97} & 78.11\% & 15.55 \\
\hline
\end{tabular}%
}
\end{center}
\end{table}

\subsubsection{Reward Ablation with Efficiency Analysis}
\paragraph{Reward ablation.}
The right part of Table~\ref{tab:ablation} studies the design of the reward. R\_$+$a is the setting used in our main experiments, where $k=1$ is kept during the whole training. We first change the value of $k$ from the start of training. As $k$ decreases from 1 to 0.5, 0 and $-1$, the number of kept edges drops from 15.83 to 12.70, 9.17 and 7.18, and the cost drops from 93.79\% to 86.91\%, 81.20\% and 77.14\%. This shows that $k$ directly controls the sparsity of the debate graph. However, the accuracy also drops to 62.33\%, 59.50\% and 60.00\%, which means that pruning too much from the start removes useful communication. We then remove the scale differences in the reward step by step. In w/o Round-Var., all rounds use the same scale $a$, while our main setting uses $a=1$ in the first round and $a=1.5$ in the second round. The accuracy drops from 65.17\% to 64.67\%. Vanilla $\pm1$ further removes the scale differences within a round and only gives rewards of $+1$ or $-1$, which lowers the accuracy to 64.33\%. The steady drop shows that both the larger weight on later rounds and the different reward values within a round contribute to the performance. Giving later rounds a larger weight helps the model focus on actions that are closer to the final result, and the different reward values within a round help the model tell apart actions that fix a wrong answer, break a correct answer or bring no change. We also compare our reward with other ways to control sparsity. Edge-Reg adds an explicit regularization on edge logits to encourage sparsity. It reaches 63.33\% with a cost of 90.56\%, which is lower than CPP in accuracy. In contrast, our reward controls sparsity directly through $k$ without any extra regularization term. For example, Second\_0 reaches a higher accuracy of 64.00\% with a lower cost of 89.00\% than Edge-Reg. This is an advantage of our reward design, since sparsity and accuracy are balanced inside one reward instead of being handled by two separate terms. The reward of MasHost, which is also designed for debate, only reaches 61.67\%, which is 3.5 points lower than ours. This shows that a reward tailored for multi-round debate is necessary. Finally, we study two-stage training. The Second rows first train with $k=1$ and then change $k$ to 0.5, 0 or $-1$ in the second stage. Compared with setting the same $k$ from the start, two-stage training keeps much more accuracy. With $k=0.5$, the accuracy rises from 62.33\% to 62.83\%. With $k=0$, it rises from 59.50\% to 64.00\%, and with $k=-1$, it rises from 60.00\% to 63.00\%. At the same time, the number of edges stays close to that of CPP, while the cost drops to around 89\% to 91\%. This suggests that the first stage learns which edges are useful, and the second stage then removes costly edges that bring little benefit. Overall, these results show that our reward design is effective, and that the coefficient $k$ offers a simple and flexible way to trade accuracy for lower cost.

\begin{table}[htbp]
\caption{Ablation study on Pipeline and Reward design.}
\label{tab:ablation}
\begin{center}
\resizebox{\linewidth}{!}{%
\begin{tabular}{l cccc | l cccc}
\hline
\multicolumn{1}{c}{} & \multicolumn{4}{c}{\bf Pipeline} & \multicolumn{1}{c}{} & \multicolumn{4}{c}{\bf Reward} \\
\cline{2-5} \cline{7-10}
\multicolumn{1}{c}{\bf Ablation} &\multicolumn{1}{c}{Ref} &\multicolumn{1}{c}{Acc (\%)} &\multicolumn{1}{c}{Cost} &\multicolumn{1}{c}{Edge} & \multicolumn{1}{c}{\bf Ablation} &\multicolumn{1}{c}{Ref} &\multicolumn{1}{c}{Acc (\%)} &\multicolumn{1}{c}{Cost} &\multicolumn{1}{c}{Edge}
\\ \hline
10-Sample  & 0.91, 0.86, 0.86 & \textbf{66.00 $\pm$ 2.37} & 96.73\% & 15.77 & R\_ + a    & 0.91, 0.87, 0.86 & \textbf{65.17 $\pm$ 3.37} & 93.79\% & 15.83 \\
30-Sample  & 0.95, 0.92, 0.90 & 64.67 $\pm$ 3.08 & 97.34\% & 16.61 & R\_0.5a & 0.73, 0.7, 0.69   & 62.33 $\pm$ 2.25 & 86.91\% & 12.70 \\
90-Sample  & 0.92, 0.88, 0.87 & 64.00 $\pm$ 1.10 & 96.27\% & 16.08 & R\_0    & 0.53, 0.5, 0.5    & 59.50 $\pm$ 2.66 & 81.20\% & 9.17  \\
150-Sample & 0.87, 0.85, 0.84 & 62.33 $\pm$ 3.50 & 94.24\% & 15.34 & R\_ - a   & 0.39, 0.39, 0.41  & 60.00 $\pm$ 2.10 & 77.14\% & 7.18  \\
\hline
w/o Training  & 0.95, 0.92, 0.91 & 63.33 $\pm$ 2.94 & 96.66\% & 16.73 & w/o Round-Var. & 0.83, 0.8, 0.79  & 64.67 $\pm$ 1.86 & 92.94\% & 14.50 \\
Random        & 0.92, 0.87, 0.87 & 63.83 $\pm$ 2.79 & 95.20\% & 15.97 & Vanilla $\pm 1$              & 0.91, 0.87, 0.86 & 64.33 $\pm$ 2.80 & 92.08\% & 15.84 \\
Text-Profile  & 0.76, 0.74, 0.73 & 61.67 $\pm$ 2.66 & 87.60\% & 13.36 & Edge-Reg             & 0.8, 0.76, 0.76   & 63.33 $\pm$ 3.88 & 90.56\% & 13.92 \\
Symbolic-MoE  & 0.86, 0.84, 0.81 & 60.00 $\pm$ 3.46 & 94.16\% & 15.07 & MasHost              & 0.85, 0.8, 0.82   & 61.67 $\pm$ 3.56 & 89.76\% & 14.85 \\
\hline
Query-Linear  & 0.91, 0.87, 0.86 & 63.67 $\pm$ 1.75 & 92.23\% & 15.85 & Second\_{0.5a}  & 0.89, 0.85, 0.85 & 62.83 $\pm$ 2.40 & 91.31\% & 15.57 \\
BE-Dominated  & 0.99, 0.99, 0.98 & 64.17 $\pm$ 2.79 & 99.89\% & 17.78 & Second\_{0}     & 0.9, 0.86, 0.86  & 64.00 $\pm$ 3.69 & 89.00\% & 15.77 \\
w/o BE        & 0.28, 0.31, 0.34 & 57.83 $\pm$ 4.26 & 80.87\% & 5.53  & Second\_- a    & 0.87, 0.83, 0.82  & 63.00 $\pm$ 2.61 & 91.47\% & 15.12 \\
\hline
\end{tabular}%
}
\end{center}
\end{table}


\begin{table}[htbp]
\caption{\textbf{Accuracy (\%, mean $\pm$ std)} on the left and \textbf{Cost} on the right. For the first 3 lines, the columns denote the token limit configurations.}
\label{tab:efficiency}
\begin{center}
\resizebox{\linewidth}{!}{%
\begin{tabular}{l cccc}
\hline
\multicolumn{1}{c}{\bf Method} &\multicolumn{1}{c}{q5/ q5/ q5} &\multicolumn{1}{c}{q75/ q75/ q75} &\multicolumn{1}{c}{q80/ q90/ q90} &\multicolumn{1}{c}{q85/ q95/ q95}
\\ \hline
Full-Debate  & 52.83 $\pm$ 3.31 \quad 68.52\% & 58.83 $\pm$ 2.04 \quad 85.87\% & 60.00 $\pm$ 2.00 \quad 88.51\% & 60.67 $\pm$ 2.94 \quad 90.30\% \\
ARG-Designer & 52.67 $\pm$ 2.73 \quad 68.24\% & 61.83 $\pm$ 4.12 \quad 85.26\% & 59.00 $\pm$ 2.37 \quad 87.58\% & 60.33 $\pm$ 3.14 \quad 89.92\% \\
Adv\_CPP     & \textbf{53.83 $\pm$ 2.04 \quad 66.26\%} & 57.00 $\pm$ 3.52 \quad 80.77\% & \textbf{62.17 $\pm$ 3.54 \quad 83.27\%} & \textbf{63.00 $\pm$ 3.16 \quad 84.48\%} \\
\hline
IoE-Vote$^{*}$      & \multicolumn{4}{c}{52.83 $\pm$ 3.97 \quad 78.82\%} \\
Debate-IoE          & \multicolumn{4}{c}{61.00 $\pm$ 2.53 \quad 87.39\%} \\
\hline
Consistency-Fuse-12 & \multicolumn{4}{c}{62.00 $\pm$ 1.67 \quad 68.46\%} \\
Consistency-Fuse-13 & \multicolumn{4}{c}{\textbf{62.00 $\pm$ 1.67 \quad 81.70\%}} \\
Consistency-Fuse-14 & \multicolumn{4}{c}{62.50 $\pm$ 0.84 \quad 94.94\%} \\
\hline
\end{tabular}%
}
\end{center}
\end{table}
\paragraph{Efficiency.}Furthermore, we find that pruning alone does not reduce the cost very much. Since output tokens are more expensive than input tokens, we further reduce the cost by limiting the number of output tokens. We first collect the statistics of output tokens of each LLM on the validation set. We then truncate the outputs only in the first and second rounds, which means that the agents only say part of what they would normally say. In the last round, the agents give complete outputs to obtain the final answer. Each configuration in Table~\ref{tab:efficiency} sets the token limits to the corresponding quantiles of the output token statistics. The bold results in the table are Pareto-optimal. Adv-CPP produces several Pareto-optimal results, and it is the only debate method that reaches the Pareto frontier. Under every token limit configuration, each result of Full-Debate and ARG-Designer is dominated by a result of Adv-CPP, which achieves a higher accuracy with a lower cost. For example, under q85/q95/q95, Adv-CPP reaches 63.00\% with a cost of 84.48\%, while Full-Debate and ARG-Designer only reach 60.67\% and 60.33\% with costs above 89\%. Adv-CPP also compares well with consistency methods. With the q85/q95/q95 configuration, it outperforms Consistency-Fuse-14 by 0.5 points while reducing the cost by more than 10 points. Under the loosest limit, Adv-CPP still keeps the lowest cost of 66.26\% among all debate methods. These results show that CPP can be combined with output truncation to further reduce the cost, and that its pruning decisions remain effective when the agents only give partial responses in the early rounds. All detailed output tokens statistics can be found in Appendix~\ref{mmmm}.

\subsubsection{Pipeline Ablation}
\paragraph{Pipeline ablation.}
The left part of Table~\ref{tab:ablation} studies the design of the pipeline. We first vary the number of validation samples used to initialize the behavior embedding, while our main experiments use 60 samples. Accuracy does not increase with more samples. Using only 10 samples already reaches 66.00\%, while using 30, 90 and 150 samples gives 64.67\%, 64.00\% and 62.33\%, respectively. This shows that a small number of responses is enough to capture the behavior of each agent, so our initialization is cheap to obtain. A possible reason for the drop is that too many responses make the embedding more general and weaken the specific behavior of each agent. We then study the effect of training and initialization. Without training, CPP only reaches 63.33\%, which is close to Random with 63.83\%. This shows that the gain of CPP mainly comes from reinforcement learning rather than from the initialization alone. When the behavior embedding is initialized with text-introduction profiles, the accuracy drops to 61.67\% and the model keeps fewer edges. The initialization used in Symbolic-MoE, which adds statistical information on top of profiles, performs even worse with 60.00\%. Both are clearly lower than our response-based initialization, which confirms that encoding the actual responses of agents provides richer information about their behavior than a general description. Next, we study the structure of FuseGLU. Query-Linear removes the fusion inside the GLU, which lowers the nonlinearity applied to the query. It keeps almost the same number of edges as CPP, with 15.85 compared to 15.83, but its accuracy drops to 63.67\%. This shows that the gain of FuseGLU comes from better decisions on which edges to keep, not from keeping more edges. BE-Dominated swaps the positions of the behavior embedding and the query, so that the behavior embedding leads the computation. In this case, the model keeps almost all edges, with reference rates close to 0.99 and a cost of 99.89\%, while its accuracy drops to 64.17\%. The model tends to decide based on agent identity and loses its ability to prune according to the content of the debate. Removing the behavior embedding causes the largest drop among all variants. The model keeps only 5.53 edges and its accuracy falls to 57.83\%. Without knowing which agent is making the decision, the model cannot judge when communication is useful and therefore prunes too much. Taken together, these results show that the behavior embedding is necessary, and that it works best as extra information fused into the gate and up projections rather than as the main input. They also show that the full design of CPP, including response-based initialization, reinforcement learning and the fusion structure of FuseGLU, is needed to reach the best balance between accuracy and cost.

\section{Conclusion}
We propose the Conditional Progressive Pruning framework. It is the first to prune MAD by efficiently using intermediate information between rounds and modeling the conditions across actions. This fully releases the potential of communication-based collaboration. CPP is also the first SoTA MAD framework that fully outperforms Single Agent and Heterogeneous Consistency methods under the same computational cost. Overall, we are the first to fully verify the advantage of MAD, which ends the debate on whether MAD is valuable. We also analyze the promising room for optimizing MAD and lay an empirical foundation for future multi-agent work.

\newpage
\section*{AI use statement}
When writing the code, we did not use AI to generate any new code. However, we did use AI to check our handwritten code and asked it for optimization suggestions. The final version of the code contains no part directly generated by AI. When writing the paper, most of our tables and figures were generated with the help of AI based on experimental results obtained by humans, i.e., we used AI to generate tables and figures. We did not use AI to generate any raw data or to explore scientific ideas. We also used AI to polish the English of our draft and to give writing suggestions. Finally, we did not use AI at all to make mathematical claims, provide key elements for proving mathematical claims, help write proofs, propose or refine hypotheses, design research methods or experiments or give feedback on them, or implement methods.


\bibliography{iclr2027_conference}

@article{du2023improving,
  title={Improving factuality and reasoning in language models through multiagent debate},
  author={Du, Yilun and Li, Shuang and Torralba, Antonio and Tenenbaum, Joshua B and Mordatch, Igor},
  journal={arXiv preprint arXiv:2305.14325},
  year={2023}
}

@article{liu2023dynamic,
  title={Dynamic llm-agent network: An llm-agent collaboration framework with agent team optimization},
  author={Liu, Zijun and Zhang, Yanzhe and Li, Peng and Liu, Yang and Yang, Diyi},
  year={2023}
}

@inproceedings{li2026ofa,
  title={OFA-MAS: One-for-all multi-agent system topology design based on mixture-of-experts graph generative models},
  author={Li, Shiyuan and Liu, Yixin and Zheng, Yu and Li, Mei and Nguyen, Quoc Viet Hung and Pan, Shirui},
  booktitle={Proceedings of the ACM Web Conference 2026},
  pages={1333--1344},
  year={2026}
}

@article{choi2026debate,
  title={Debate or vote: Which yields better decisions in multi-agent large language models?},
  author={Choi, Hyeong Kyu and Zhu, Jerry and Li, Sharon},
  journal={Advances in Neural Information Processing Systems},
  volume={38},
  pages={101732--101764},
  year={2026}
}

@inproceedings{huang2024large,
  title={Large language models cannot self-correct reasoning yet},
  author={Huang, Jie and Chen, Xinyun and Mishra, Swaroop and Zheng, Huaixiu Steven and Yu, Adams and Song, Xinying and Zhou, Denny},
  booktitle={International conference on learning representations},
  volume={2024},
  pages={32808--32824},
  year={2024}
}

@article{li2024more,
  title={More agents is all you need},
  author={Li, Junyou and Zhang, Qin and Yu, Yangbin and Fu, Qiang and Ye, Deheng},
  journal={arXiv preprint arXiv:2402.05120},
  year={2024}
}

@article{zhang2023exploring,
  title={Exploring collaboration mechanisms for llm agents: A social psychology view},
  author={Zhang, Jintian and Xu, Xin and Zhang, Ningyu and Liu, Ruibo and Hooi, Bryan and Deng, Shumin},
  journal={arXiv preprint arXiv:2310.02124},
  year={2023}
}

@inproceedings{chai2024expert,
  title={An expert is worth one token: Synergizing multiple expert llms as generalist via expert token routing},
  author={Chai, Ziwei and Wang, Guoyin and Su, Jing and Zhang, Tianjie and Huang, Xuanwen and Wang, Xuwu and Xu, Jingjing and Yuan, Jianbo and Yang, Hongxia and Wu, Fei and others},
  booktitle={Proceedings of the 62nd Annual Meeting of the Association for Computational Linguistics (Volume 1: Long Papers)},
  pages={11385--11396},
  year={2024}
}

@inproceedings{wang2025anymac,
  title={Anymac: Cascading flexible multi-agent collaboration via next-agent prediction},
  author={Wang, Song and Tan, Zhen and Chen, Zihan and Zhou, Shuang and Chen, Tianlong and Li, Jundong},
  booktitle={Proceedings of the 2025 Conference on Empirical Methods in Natural Language Processing},
  pages={11566--11578},
  year={2025}
}

@article{wang2022self,
  title={Self-consistency improves chain of thought reasoning in language models},
  author={Wang, Xuezhi and Wei, Jason and Schuurmans, Dale and Le, Quoc and Chi, Ed and Narang, Sharan and Chowdhery, Aakanksha and Zhou, Denny},
  journal={arXiv preprint arXiv:2203.11171},
  year={2022}
}

@article{zheng2023progressive,
  title={Progressive-hint prompting improves reasoning in large language models},
  author={Zheng, Chuanyang and Liu, Zhengying and Xie, Enze and Li, Zhenguo and Li, Yu},
  journal={arXiv preprint arXiv:2304.09797},
  year={2023}
}

@article{shinn2023reflexion,
  title={Reflexion: Language agents with verbal reinforcement learning},
  author={Shinn, Noah and Cassano, Federico and Gopinath, Ashwin and Narasimhan, Karthik and Yao, Shunyu},
  journal={Advances in neural information processing systems},
  volume={36},
  pages={8634--8652},
  year={2023}
}

@inproceedings{zeng2025s2,
  title={S2-mad: Breaking the token barrier to enhance multi-agent debate efficiency},
  author={Zeng, Yuting and Huang, Weizhe and Jiang, Lei and Liu, Tongxuan and Jin, Xitai and Tiana, Chen Tianying and Li, Jing and Xu, Xiaohua},
  booktitle={Proceedings of the 2025 Conference of the Nations of the Americas Chapter of the Association for Computational Linguistics: Human Language Technologies (Volume 1: Long Papers)},
  pages={9393--9408},
  year={2025}
}

@article{kim2024mdagents,
  title={Mdagents: An adaptive collaboration of llms for medical decision-making},
  author={Kim, Yubin and Park, Chanwoo and Jeong, Hyewon and Chan, Yik S and Xu, Xuhai and McDuff, Daniel and Lee, Hyeonhoon and Ghassemi, Marzyeh and Breazeal, Cynthia and Park, Hae W},
  journal={Advances in Neural Information Processing Systems},
  volume={37},
  pages={79410--79452},
  year={2024}
}

@article{zhuge2024language,
  title={Language agents as optimizable graphs},
  author={Zhuge, Mingchen and Wang, Wenyi and Kirsch, Louis and Faccio, Francesco and Khizbullin, Dmitrii and Schmidhuber, J{\"u}rgen},
  journal={arXiv preprint arXiv:2402.16823},
  year={2024}
}

@inproceedings{zhang2025cut,
  title={Cut the crap: An economical communication pipeline for llm-based multi-agent systems},
  author={Zhang, Guibin and Yue, Yanwei and Li, Zhixun and Yun, Sukwon and Wan, Guancheng and Wang, Kun and Cheng, Dawei and Yu, Jeffrey and Chen, Tianlong},
  booktitle={International Conference on Learning Representations},
  volume={2025},
  pages={75389--75428},
  year={2025}
}

@inproceedings{wang2025agentdropout,
  title={Agentdropout: Dynamic agent elimination for token-efficient and high-performance llm-based multi-agent collaboration},
  author={Wang, Zhexuan and Wang, Yutong and Liu, Xuebo and Ding, Liang and Zhang, Miao and Liu, Jie and Zhang, Min},
  booktitle={Proceedings of the 63rd Annual Meeting of the Association for Computational Linguistics (Volume 1: Long Papers)},
  pages={24013--24035},
  year={2025}
}

@article{zhang2024g,
  title={G-designer: Architecting multi-agent communication topologies via graph neural networks},
  author={Zhang, Guibin and Yue, Yanwei and Sun, Xiangguo and Wan, Guancheng and Yu, Miao and Fang, Junfeng and Wang, Kun and Chen, Tianlong and Cheng, Dawei},
  journal={arXiv preprint arXiv:2410.11782},
  year={2024}
}

@article{ye2025mas,
  title={Mas-gpt: Training llms to build llm-based multi-agent systems},
  author={Ye, Rui and Tang, Shuo and Ge, Rui and Du, Yaxin and Yin, Zhenfei and Chen, Siheng and Shao, Jing},
  journal={arXiv preprint arXiv:2503.03686},
  year={2025}
}

@inproceedings{li2026assemble,
  title={Assemble your crew: Automatic multi-agent communication topology design via autoregressive graph generation},
  author={Li, Shiyuan and Liu, Yixin and Wen, Qingsong and Zhang, Chengqi and Pan, Shirui},
  booktitle={Proceedings of the AAAI Conference on Artificial Intelligence},
  volume={40},
  number={28},
  pages={23142--23150},
  year={2026}
}

@article{yang2025mashost,
  title={MasHost builds it all: autonomous multi-agent system directed by reinforcement learning},
  author={Yang, Kuo and Yang, Xingjie and Yu, Linhui and Xu, Qing and Fang, Yan and Wang, Xu and Zhou, Zhengyang and Wang, Yang},
  journal={arXiv preprint arXiv:2506.08507},
  year={2025}
}

@article{chen2025symbolic,
  title={Symbolic mixture-of-experts: Adaptive skill-based routing for heterogeneous reasoning},
  author={Chen, Justin and Yun, Sukwon and Stengel-Eskin, Elias and Chen, Tianlong and Bansal, Mohit},
  year={2025}
}

@article{hendrycks2021measuring,
  title={Measuring mathematical problem solving with the math dataset},
  author={Hendrycks, Dan and Burns, Collin and Kadavath, Saurav and Arora, Akul and Basart, Steven and Tang, Eric and Song, Dawn and Steinhardt, Jacob},
  journal={arXiv preprint arXiv:2103.03874},
  year={2021}
}

@article{wang2024mmlu,
  title={Mmlu-pro: A more robust and challenging multi-task language understanding benchmark},
  author={Wang, Yubo and Ma, Xueguang and Zhang, Ge and Ni, Yuansheng and Chandra, Abhranil and Guo, Shiguang and Ren, Weiming and Arulraj, Aaran and He, Xuan and Jiang, Ziyan and others},
  journal={Advances in Neural Information Processing Systems},
  volume={37},
  pages={95266--95290},
  year={2024}
}

@article{rein2023gpqa,
  title={Gpqa: A graduate-level google-proof q\&a benchmark},
  author={Rein, David and Hou, Betty Li and Stickland, Asa Cooper and Petty, Jackson and Pang, Richard Yuanzhe and Dirani, Julien and Michael, Julian and Bowman, Samuel R},
  journal={arXiv preprint arXiv:2311.12022},
  year={2023}
}

@article{yang2025qwen3,
  title={Qwen3 technical report},
  author={Yang, An and Li, Anfeng and Yang, Baosong and Zhang, Beichen and Hui, Binyuan and Zheng, Bo and Yu, Bowen and Gao, Chang and Huang, Chengen and Lv, Chenxu and others},
  journal={arXiv preprint arXiv:2505.09388},
  year={2025}
}

@article{li2024confidence,
  title={Confidence matters: Revisiting intrinsic self-correction capabilities of large language models},
  author={Li, Loka and Chen, Zhenhao and Chen, Guangyi and Zhang, Yixuan and Su, Yusheng and Xing, Eric and Zhang, Kun},
  journal={arXiv preprint arXiv:2402.12563},
  year={2024}
}
\bibliographystyle{iclr2027_conference}

\newpage
\appendix

\section{Action-Wise Influence Dependency}
\label{AAA}

\begin{table}[htbp]
\caption{Dependency structure of $A_1$ and $A_2$ under $3 \times 3$ MAD Architecture.}
\label{tab:action-dependency}
\begin{center}
\begin{tabular}{l l}
\hline
\bf Action & \bf Dependency \\
\hline
$A_1[0][0]$ & \begin{tabular}[t]{@{}l@{\quad}l@{\quad}l@{\quad}l@{}}
$(A_1[0][0], 1)$ & $(A_1[0][1], \gamma)$ & $(A_1[0][2], \gamma^2)$ & $(A_2[0][0], \beta)$ \\
$(A_2[0][1], \beta\gamma)$ & $(A_2[0][2], \beta\gamma^2)$ & $(A_2[1][0], \beta\gamma)$ & $(A_2[1][2], \beta\gamma^2)$ \\
$(A_2[2][0], \beta\gamma)$ & $(A_2[2][1], \beta\gamma^2)$ &  &  \\
\end{tabular} \\
\hline
$A_1[0][1]$ & \begin{tabular}[t]{@{}l@{\quad}l@{\quad}l@{\quad}l@{}}
$(A_1[0][1], 1)$ & $(A_1[0][2], \gamma)$ & $(A_2[0][0], \beta)$ & $(A_2[0][1], \beta\gamma)$ \\
$(A_2[0][2], \beta\gamma^2)$ & $(A_2[1][0], \beta\gamma)$ & $(A_2[1][2], \beta\gamma^2)$ & $(A_2[2][0], \beta\gamma)$ \\
$(A_2[2][1], \beta\gamma^2)$ &  &  &  \\
\end{tabular} \\
\hline
$A_1[0][2]$ & \begin{tabular}[t]{@{}l@{\quad}l@{\quad}l@{\quad}l@{}}
$(A_1[0][2], 1)$ & $(A_2[0][0], \beta)$ & $(A_2[0][1], \beta\gamma)$ & $(A_2[0][2], \beta\gamma^2)$ \\
$(A_2[1][0], \beta\gamma)$ & $(A_2[1][2], \beta\gamma^2)$ & $(A_2[2][0], \beta\gamma)$ & $(A_2[2][1], \beta\gamma^2)$ \\
\end{tabular} \\
\hline
$A_1[1][0]$ & \begin{tabular}[t]{@{}l@{\quad}l@{\quad}l@{\quad}l@{}}
$(A_1[1][0], 1)$ & $(A_1[1][2], \gamma)$ & $(A_2[0][1], \beta\gamma)$ & $(A_2[0][2], \beta\gamma^2)$ \\
$(A_2[1][1], \beta)$ & $(A_2[1][0], \beta\gamma)$ & $(A_2[1][2], \beta\gamma^2)$ & $(A_2[2][1], \beta\gamma^2)$ \\
\end{tabular} \\
\hline
$A_1[1][1]$ & \begin{tabular}[t]{@{}l@{\quad}l@{\quad}l@{\quad}l@{}}
$(A_1[1][1], 1)$ & $(A_1[1][0], \gamma)$ & $(A_1[1][2], \gamma^2)$ & $(A_2[0][1], \beta\gamma)$ \\
$(A_2[0][2], \beta\gamma^2)$ & $(A_2[1][1], \beta)$ & $(A_2[1][0], \beta\gamma)$ & $(A_2[1][2], \beta\gamma^2)$ \\
$(A_2[2][1], \beta\gamma^2)$ &  &  &  \\
\end{tabular} \\
\hline
$A_1[1][2]$ & \begin{tabular}[t]{@{}l@{\quad}l@{\quad}l@{\quad}l@{}}
$(A_1[1][2], 1)$ & $(A_2[0][1], \beta\gamma)$ & $(A_2[0][2], \beta\gamma^2)$ & $(A_2[1][1], \beta)$ \\
$(A_2[1][0], \beta\gamma)$ & $(A_2[1][2], \beta\gamma^2)$ & $(A_2[2][1], \beta\gamma^2)$ &  \\
\end{tabular} \\
\hline
$A_1[2][0]$ & \begin{tabular}[t]{@{}l@{\quad}l@{\quad}l@{\quad}l@{}}
$(A_1[2][0], 1)$ & $(A_1[2][1], \gamma)$ & $(A_2[0][2], \beta\gamma^2)$ & $(A_2[1][2], \beta\gamma^2)$ \\
$(A_2[2][2], \beta)$ & $(A_2[2][0], \beta\gamma)$ & $(A_2[2][1], \beta\gamma^2)$ &  \\
\end{tabular} \\
\hline
$A_1[2][1]$ & \begin{tabular}[t]{@{}l@{\quad}l@{\quad}l@{\quad}l@{}}
$(A_1[2][1], 1)$ & $(A_2[0][2], \beta\gamma^2)$ & $(A_2[1][2], \beta\gamma^2)$ & $(A_2[2][2], \beta)$ \\
$(A_2[2][0], \beta\gamma)$ & $(A_2[2][1], \beta\gamma^2)$ &  &  \\
\end{tabular} \\
\hline
$A_1[2][2]$ & \begin{tabular}[t]{@{}l@{\quad}l@{\quad}l@{\quad}l@{}}
$(A_1[2][2], 1)$ & $(A_1[2][0], \gamma)$ & $(A_1[2][1], \gamma^2)$ & $(A_2[0][2], \beta\gamma^2)$ \\
$(A_2[1][2], \beta\gamma^2)$ & $(A_2[2][2], \beta)$ & $(A_2[2][0], \beta\gamma)$ & $(A_2[2][1], \beta\gamma^2)$ \\
\end{tabular} \\
\hline
$A_2[0][0]$ & \begin{tabular}[t]{@{}l@{\quad}l@{\quad}l@{\quad}l@{}}
$(A_2[0][0], 1)$ & $(A_2[0][1], \gamma)$ & $(A_2[0][2], \gamma^2)$ &  \\
\end{tabular} \\
\hline
$A_2[0][1]$ & \begin{tabular}[t]{@{}l@{\quad}l@{\quad}l@{\quad}l@{}}
$(A_2[0][1], 1)$ & $(A_2[0][2], \gamma)$ &  &  \\
\end{tabular} \\
\hline
$A_2[0][2]$ & \begin{tabular}[t]{@{}l@{\quad}l@{\quad}l@{\quad}l@{}}
$(A_2[0][2], 1)$ &  &  &  \\
\end{tabular} \\
\hline
$A_2[1][0]$ & \begin{tabular}[t]{@{}l@{\quad}l@{\quad}l@{\quad}l@{}}
$(A_2[1][0], 1)$ & $(A_2[1][2], \gamma)$ &  &  \\
\end{tabular} \\
\hline
$A_2[1][1]$ & \begin{tabular}[t]{@{}l@{\quad}l@{\quad}l@{\quad}l@{}}
$(A_2[1][1], 1)$ & $(A_2[1][0], \gamma)$ & $(A_2[1][2], \gamma^2)$ &  \\
\end{tabular} \\
\hline
$A_2[1][2]$ & \begin{tabular}[t]{@{}l@{\quad}l@{\quad}l@{\quad}l@{}}
$(A_2[1][2], 1)$ &  &  &  \\
\end{tabular} \\
\hline
$A_2[2][0]$ & \begin{tabular}[t]{@{}l@{\quad}l@{\quad}l@{\quad}l@{}}
$(A_2[2][0], 1)$ & $(A_2[2][1], \gamma)$ &  &  \\
\end{tabular} \\
\hline
$A_2[2][1]$ & \begin{tabular}[t]{@{}l@{\quad}l@{\quad}l@{\quad}l@{}}
$(A_2[2][1], 1)$ &  &  &  \\
\end{tabular} \\
\hline
$A_2[2][2]$ & \begin{tabular}[t]{@{}l@{\quad}l@{\quad}l@{\quad}l@{}}
$(A_2[2][2], 1)$ & $(A_2[2][0], \gamma)$ & $(A_2[2][1], \gamma^2)$ &  \\
\end{tabular} \\
\hline
\end{tabular}
\end{center}
\end{table}

\section{Edge Statistics of Table~\ref{tab:main}}
\label{abcde}

\begin{table}[htbp]
\caption{Comparison of \textbf{Referenced Ratio} (Grok, GPT, Mistral) and \textbf{Edge Number} across MATH-Hard, GPQA-Diamond and MMLU-Pro.}
\label{tab:mainedge}
\begin{center}
\resizebox{\linewidth}{!}{%
\begin{tabular}{lcc|cc|cc}
\hline
\multicolumn{1}{c}{} & \multicolumn{2}{c}{\bf MATH-Hard} & \multicolumn{2}{c}{\bf GPQA-Diamond} & \multicolumn{2}{c}{\bf MMLU-Pro} \\
\cline{2-3} \cline{4-5} \cline{6-7}
\multicolumn{1}{c}{\bf Method} &\multicolumn{1}{c}{Ref} &\multicolumn{1}{c}{Edge} &\multicolumn{1}{c}{Ref} &\multicolumn{1}{c}{Edge} &\multicolumn{1}{c}{Ref} &\multicolumn{1}{c}{Edge}
\\ \hline
Full-Debate   & 1, 1, 1          & 18.00 & 1, 1, 1          & 18.00 & 1, 1, 1          & 18.00 \\
Random        & 0.5, 0.51, 0.5   & 9.02  & 0.5, 0.5, 0.5    & 9.02  & 0.5, 0.5, 0.5    & 9.02  \\
LLM-Self      & 1.67, 1.58, 1.62 & 29.24 & 1.76, 1.47, 1.51 & 28.48 & 1.77, 1.58, 1.61 & 29.80 \\
Intervention  & 0.4, 0.77, 0.83  & 12.00 & 0.53, 0.75, 0.69 & 12.00 & 0.56, 0.72, 0.72 & 12.00 \\
$S^2$-MAD     & 0.57, 0.58, 0.57 & 10.31 & 0.52, 0.56, 0.55 & 9.77  & 0.45, 0.47, 0.46 & 8.24  \\
IoE-Debate    & 0.67, 0.67, 0.67 & 12.00 & 0.67, 0.67, 0.67 & 12.00 & 0.67, 0.67, 0.67 & 12.00 \\
Debate-IoE    & 0.67, 0.67, 0.67 & 12.00 & 0.67, 0.67, 0.67 & 12.00 & 0.67, 0.67, 0.67 & 12.00 \\
Zero          & 0, 0, 0          & 0.00  & 0, 0, 0          & 0.00  & 0, 0, 0          & 0.00  \\
\hline
ARG-Designer  & 0.97, 0.95, 0.96 & 17.32 & 0.68, 0.66, 0.69 & 12.15 & 0.28, 0.22, 0.23 & 4.43  \\
AnyMac        & 0.8, 0.8, 0.79   & 14.28 & 0.7, 0.67, 0.68  & 12.22 & 0.6, 0.58, 0.57  & 10.55 \\ CPP  & 0.85, 0.84, 0.84 & 15.24 & 0.82, 0.78, 0.8  & 14.31 & 0.98, 0.97, 0.97 & 17.52 \\
\hline
\end{tabular}%
}
\end{center}
\end{table}

\section{Output Token Statistics on MATH-Hard}
\label{mmmm}


\begin{table}[htbp]
    \centering
    \caption{Output Token limit configurations at different quantiles on MATH-Hard Validation}
    \label{tab:limits}
    \begin{tabular}{lccc}
        \toprule
        Threshold & Grok & GPT & Mistral \\
        \midrule
        q5   & 65   & 132  & 139  \\
        q75  & 157  & 318  & 432  \\
        q80  & 166  & 353  & 478  \\
        q85  & 177  & 393  & 527  \\
        q90  & 196  & 458  & 623  \\
        q95  & 233  & 621  & 872  \\
        q100 & 6000 & 6000 & 6000 \\
        \bottomrule
    \end{tabular}
\end{table}

\section{Detailed Consistency Results}
\label{dcons}

\begin{table}[htbp]
\caption{Consistency Accuracy (\%) of Grok across different sample sizes $n$ on MATH-Hard-I.}
\label{tab:grokM1-n-sweep}
\begin{center}
\resizebox{\linewidth}{!}{%
\begin{tabular}{c|cccccccc|c}
\hline
\bf $n$ & 2 & 3 & 4 & 5 & 6 & 7 & 8 & 9 & \bf Cost \\
\hline
Grok & \makecell{49.33\\$\pm$2.34} & \makecell{51.17\\$\pm$2.40} & \makecell{53.33\\$\pm$3.83} & \makecell{55.00\\$\pm$3.10} & \makecell{56.00\\$\pm$0.63} & \makecell{56.83\\$\pm$1.94} & \makecell{57.83\\$\pm$1.47} & \makecell{58.50\\$\pm$1.97} & 101.06\% \\
\hline
\end{tabular}%
}
\end{center}
\end{table}

\begin{table}[htbp]
\caption{Consistency Accuracy (\%) of GPT across different sample sizes $n$ on MATH-Hard-I.}
\label{tab:gptM1-n-sweep}
\begin{center}
\resizebox{\linewidth}{!}{%
\begin{tabular}{c|cccccccccccc|c}
\hline
\bf $n$ & 2 & 3 & 4 & 5 & 6 & 7 & 8 & 9 & 10 & 11 & 12 & 13 & \bf Cost \\
\hline
GPT & \makecell{47.67\\$\pm$2.25} & \makecell{49.33\\$\pm$2.42} & \makecell{51.67\\$\pm$1.51} & \makecell{53.33\\$\pm$2.07} & \makecell{53.50\\$\pm$2.43} & \makecell{54.00\\$\pm$1.79} & \makecell{54.67\\$\pm$1.03} & \makecell{55.83\\$\pm$0.98} & \makecell{57.33\\$\pm$2.34} & \makecell{57.67\\$\pm$1.03} & \makecell{57.00\\$\pm$1.55} & \makecell{57.17\\$\pm$2.04} & 100.32\% \\
\hline
\end{tabular}%
}
\end{center}
\end{table}

\begin{table}[htbp]
\caption{Consistency Accuracy (\%) of Mistral across different sample sizes $n$ on MATH-Hard-I.}
\label{tab:mistralM1-n-sweep}
\begin{center}
\resizebox{\linewidth}{!}{%
\begin{tabular}{c|cccccccccccc|c}
\hline
\bf $n$ & 2 & 3 & 4 & 5 & 6 & 7 & 8 & 9 & 10 & 11 & 12 & 13 & \bf Cost \\
\hline
Mistral & \makecell{49.00\\$\pm$4.24} & \makecell{52.67\\$\pm$2.16} & \makecell{54.50\\$\pm$3.78} & \makecell{56.00\\$\pm$2.83} & \makecell{55.83\\$\pm$2.32} & \makecell{56.67\\$\pm$2.25} & \makecell{56.33\\$\pm$2.42} & \makecell{56.67\\$\pm$1.97} & \makecell{57.17\\$\pm$1.83} & \makecell{56.83\\$\pm$1.72} & \makecell{57.50\\$\pm$1.97} & \makecell{57.17\\$\pm$1.94} & 107.48\% \\
\hline
\end{tabular}%
}
\end{center}
\end{table}

\begin{table}[htbp]
\caption{Consistency Accuracy (\%) of Fuse-Extension across different sample sizes $n$  on MATH-Hard-I.}
\label{tab:fuseM1-extension-n-sweep}
\begin{center}
\resizebox{\linewidth}{!}{%
\begin{tabular}{c|ccccccccccc|c}
\hline
\bf $n$ & 2 & 3 & 4 & 5 & 6 & 7 & 8 & 9 & 10 & 11 & 12 & \bf Cost \\
\hline
Fuse-Extension & \makecell{49.83\\$\pm$1.83} & \makecell{52.67\\$\pm$1.21} & \makecell{53.83\\$\pm$2.64} & \makecell{55.50\\$\pm$2.43} & \makecell{58.67\\$\pm$1.03} & \makecell{59.83\\$\pm$1.17} & \makecell{60.50\\$\pm$2.59} & \makecell{62.33\\$\pm$1.63} & \makecell{62.83\\$\pm$2.14} & \makecell{63.50\\$\pm$3.27} & \makecell{64.00\\$\pm$2.45} & 107.89\% \\
\hline
\end{tabular}%
}
\end{center}
\end{table}

\begin{table}[htbp]
\caption{Consistency Accuracy (\%) of Best-Extension across different sample sizes $n$  on MATH-Hard-I.}
\label{tab:bestM1-extension-n-sweep}
\begin{center}
\resizebox{\linewidth}{!}{%
\begin{tabular}{c|cccccccccc|c}
\hline
\bf $n$ & 2 & 3 & 4 & 5 & 6 & 7 & 8 & 9 & 10 & 11 & \bf Cost \\
\hline
Best-Extension & \makecell{50.83\\$\pm$2.32} & \makecell{53.17\\$\pm$1.60} & \makecell{55.00\\$\pm$2.10} & \makecell{57.33\\$\pm$2.66} & \makecell{58.17\\$\pm$2.71} & \makecell{59.17\\$\pm$2.64} & \makecell{61.33\\$\pm$1.51} & \makecell{62.00\\$\pm$2.10} & \makecell{62.50\\$\pm$1.52} & \makecell{62.83\\$\pm$2.14} & 103.48\% \\
\hline
\end{tabular}%
}
\end{center}
\end{table}

\begin{table}[htbp]
\caption{Consistency Accuracy (\%) of Mistral across different sample sizes $n$ on MATH-Hard-II.}
\label{tab:mistralM2-n-sweep-5}
\begin{center}
\resizebox{\linewidth}{!}{%
\begin{tabular}{c|ccccccccccc}
\hline
\bf $n$ & 2 & 3 & 4 & 5 & 6 & 7 & 8 & 9 & 10 & 11 & 12 \\
\hline
Mistral & \makecell{48.50\\$\pm$4.85} & \makecell{51.83\\$\pm$3.37} & \makecell{54.33\\$\pm$3.14} & \makecell{56.33\\$\pm$2.80} & \makecell{55.50\\$\pm$1.97} & \makecell{56.50\\$\pm$2.07} & \makecell{56.67\\$\pm$2.42} & \makecell{57.00\\$\pm$2.00} & \makecell{57.00\\$\pm$2.45} & \makecell{56.83\\$\pm$1.17} & \makecell{58.00\\$\pm$1.79} \\
\hline
\end{tabular}%
}
\vspace{4pt}

\resizebox{\linewidth}{!}{%
\begin{tabular}{c|cccccccccc|c}
\hline
\bf $n$ & 13 & 14 & 15 & 16 & 17 & 18 & 19 & 20 & 21 & 22 & \bf Cost \\
\hline
Mistral & \makecell{57.00\\$\pm$2.10} & \makecell{56.67\\$\pm$2.25} & \makecell{58.17\\$\pm$3.49} & \makecell{58.83\\$\pm$3.66} & \makecell{59.00\\$\pm$3.10} & \makecell{59.00\\$\pm$3.22} & \makecell{58.17\\$\pm$2.32} & \makecell{58.67\\$\pm$2.42} & \makecell{59.50\\$\pm$2.51} & \makecell{59.67\\$\pm$2.25} & 94.62\% \\
\hline
\end{tabular}%
}
\end{center}
\end{table}

\begin{table}[htbp]
\caption{Consistency Accuracy (\%) of Fuse-Extension across different sample sizes $n$ on MATH-Hard-II.}
\label{tab:fuseM2-extension-n-sweep-5}
\begin{center}
\resizebox{\linewidth}{!}{%
\begin{tabular}{c|ccccccccccccc|c}
\hline
\bf $n$ & 2 & 3 & 4 & 5 & 6 & 7 & 8 & 9 & 10 & 11 & 12 & 13 & 14 & \bf Cost \\
\hline
Fuse-Extension & \makecell{48.00\\$\pm$3.35} & \makecell{51.17\\$\pm$2.14} & \makecell{53.33\\$\pm$2.73} & \makecell{53.33\\$\pm$2.58} & \makecell{56.50\\$\pm$2.26} & \makecell{58.33\\$\pm$2.34} & \makecell{59.17\\$\pm$0.98} & \makecell{60.00\\$\pm$1.10} & \makecell{61.83\\$\pm$1.47} & \makecell{62.17\\$\pm$1.17} & \makecell{62.00\\$\pm$1.67} & \makecell{62.00\\$\pm$1.67} & \makecell{62.50\\$\pm$0.84} & 94.94\% \\
\hline
\end{tabular}%
}
\end{center}
\end{table}

\begin{table}[htbp]
\caption{Consistency Accuracy (\%) of Best-Extension across different sample sizes $n$ on MATH-Hard-II.}
\label{tab:bestM2-extension-n-sweep-5}
\begin{center}
\resizebox{\linewidth}{!}{%
\begin{tabular}{c|ccccccccccccccc|c}
\hline
\bf $n$ & 2 & 3 & 4 & 5 & 6 & 7 & 8 & 9 & 10 & 11 & 12 & 13 & 14 & 15 & 16 & \bf Cost \\
\hline
Best-Extension & \makecell{48.00\\$\pm$3.90} & \makecell{51.00\\$\pm$2.37} & \makecell{53.33\\$\pm$3.98} & \makecell{53.17\\$\pm$2.99} & \makecell{55.83\\$\pm$2.14} & \makecell{56.50\\$\pm$2.66} & \makecell{58.17\\$\pm$2.79} & \makecell{57.33\\$\pm$3.44} & \makecell{58.50\\$\pm$1.97} & \makecell{60.00\\$\pm$1.10} & \makecell{60.50\\$\pm$1.64} & \makecell{60.33\\$\pm$1.37} & \makecell{60.83\\$\pm$1.60} & \makecell{61.50\\$\pm$1.52} & \makecell{62.17\\$\pm$1.72} & 94.81\% \\
\hline
\end{tabular}%
}
\end{center}
\end{table}

\begin{table}[htbp]
\caption{Adversarial Consistency Accuracy (\%) of Mistral across different sample sizes $n$ on MATH-Hard-II.}
\label{tab:mistralAD-n-sweep-4}
\begin{center}
\resizebox{\linewidth}{!}{%
\begin{tabular}{c|ccccccccccc}
\hline
\bf $n$ & 2 & 3 & 4 & 5 & 6 & 7 & 8 & 9 & 10 & 11 & 12 \\
\hline
Mistral & \makecell{11.67\\$\pm$4.03} & \makecell{10.50\\$\pm$1.52} & \makecell{13.50\\$\pm$2.74} & \makecell{15.17\\$\pm$1.60} & \makecell{15.67\\$\pm$3.14} & \makecell{15.50\\$\pm$1.87} & \makecell{17.00\\$\pm$2.28} & \makecell{16.67\\$\pm$2.73} & \makecell{17.83\\$\pm$1.83} & \makecell{18.00\\$\pm$2.76} & \makecell{19.17\\$\pm$1.17} \\
\hline
\end{tabular}%
}
\vspace{4pt}

\resizebox{\linewidth}{!}{%
\begin{tabular}{c|cccccccccc|c}
\hline
\bf $n$ & 13 & 14 & 15 & 16 & 17 & 18 & 19 & 20 & 21 & 22 & \bf Cost \\
\hline
Mistral & \makecell{19.67\\$\pm$1.21} & \makecell{20.83\\$\pm$1.47} & \makecell{20.50\\$\pm$1.52} & \makecell{20.50\\$\pm$3.08} & \makecell{19.50\\$\pm$2.07} & \makecell{19.50\\$\pm$2.66} & \makecell{19.17\\$\pm$2.04} & \makecell{19.00\\$\pm$1.41} & \makecell{20.50\\$\pm$2.35} & \makecell{19.83\\$\pm$1.83} & 25.79\% \\
\hline
\end{tabular}%
}
\end{center}
\end{table}

\begin{table}[htbp]
\caption{Adversarial Consistency Accuracy (\%) of Fuse-Extension across different sample sizes $n$ on MATH-Hard-II.}
\label{tab:fuseAD-extension-n-sweep-4}
\begin{center}
\resizebox{\linewidth}{!}{%
\begin{tabular}{c|ccccccccccccc|c}
\hline
\bf $n$ & 2 & 3 & 4 & 5 & 6 & 7 & 8 & 9 & 10 & 11 & 12 & 13 & 14 & \bf Cost \\
\hline
Fuse-Extension & \makecell{12.67\\$\pm$2.42} & \makecell{12.50\\$\pm$2.59} & \makecell{12.17\\$\pm$2.79} & \makecell{13.83\\$\pm$2.64} & \makecell{14.00\\$\pm$2.68} & \makecell{14.83\\$\pm$2.64} & \makecell{14.33\\$\pm$2.34} & \makecell{14.17\\$\pm$2.48} & \makecell{14.00\\$\pm$1.41} & \makecell{14.67\\$\pm$2.50} & \makecell{16.17\\$\pm$1.72} & \makecell{16.33\\$\pm$2.80} & \makecell{18.00\\$\pm$1.26} & 54.67\% \\
\hline
\end{tabular}%
}
\end{center}
\end{table}

\begin{table}[htbp]
\caption{Adversarial Consistency Accuracy (\%) of Best-Extension across different sample sizes $n$ on MATH-Hard-II.}
\label{tab:bestAD-extension-n-sweep-4}
\begin{center}
\resizebox{\linewidth}{!}{%
\begin{tabular}{c|ccccccccccccccc|c}
\hline
\bf $n$ & 2 & 3 & 4 & 5 & 6 & 7 & 8 & 9 & 10 & 11 & 12 & 13 & 14 & 15 & 16 & \bf Cost \\
\hline
Best-Extension & \makecell{12.83\\$\pm$2.56} & \makecell{12.83\\$\pm$1.47} & \makecell{13.17\\$\pm$1.94} & \makecell{15.50\\$\pm$3.62} & \makecell{14.17\\$\pm$3.54} & \makecell{15.00\\$\pm$2.45} & \makecell{15.83\\$\pm$2.32} & \makecell{16.17\\$\pm$2.14} & \makecell{16.50\\$\pm$1.87} & \makecell{16.50\\$\pm$1.22} & \makecell{16.83\\$\pm$0.98} & \makecell{17.33\\$\pm$1.75} & \makecell{16.83\\$\pm$0.75} & \makecell{18.17\\$\pm$1.60} & \makecell{17.83\\$\pm$1.94} & 47.54\% \\
\hline
\end{tabular}%
}
\end{center}
\end{table}

\begin{table}[htbp]
\caption{Consistency Accuracy (\%) of Grok across different sample sizes $n$ on GPQA-Diamond.}
\label{tab:grokG-n-sweep-2}
\begin{center}
\resizebox{\linewidth}{!}{%
\begin{tabular}{c|cccccccc|c}
\hline
\bf $n$ & 2 & 3 & 4 & 5 & 6 & 7 & 8 & 9 & \bf Cost \\
\hline
Grok & \makecell{58.00\\$\pm$2.28} & \makecell{59.83\\$\pm$2.86} & \makecell{60.33\\$\pm$2.25} & \makecell{60.83\\$\pm$2.32} & \makecell{62.50\\$\pm$2.59} & \makecell{63.33\\$\pm$1.63} & \makecell{63.17\\$\pm$1.94} & \makecell{63.00\\$\pm$1.67} & 117.55\% \\
\hline
\end{tabular}%
}
\end{center}
\end{table}

\begin{table}[htbp]
\caption{Consistency Accuracy (\%) of GPT across different sample sizes $n$ on GPQA-Diamond.}
\label{tab:gptG-n-sweep-2}
\begin{center}
\resizebox{\linewidth}{!}{%
\begin{tabular}{c|cccccccccccccc|c}
\hline
\bf $n$ & 2 & 3 & 4 & 5 & 6 & 7 & 8 & 9 & 10 & 11 & 12 & 13 & 14 & 15 & \bf Cost \\
\hline
GPT & \makecell{39.33\\$\pm$3.14} & \makecell{42.83\\$\pm$1.72} & \makecell{44.00\\$\pm$1.41} & \makecell{45.83\\$\pm$3.19} & \makecell{47.33\\$\pm$1.75} & \makecell{46.33\\$\pm$2.73} & \makecell{46.83\\$\pm$3.60} & \makecell{46.33\\$\pm$2.73} & \makecell{47.83\\$\pm$1.60} & \makecell{47.00\\$\pm$2.10} & \makecell{46.00\\$\pm$2.37} & \makecell{46.67\\$\pm$2.42} & \makecell{46.83\\$\pm$2.64} & \makecell{46.33\\$\pm$2.42} & 102.71\% \\
\hline
\end{tabular}%
}
\end{center}
\end{table}

\begin{table}[htbp]
\caption{Consistency Accuracy (\%) of Mistral across different sample sizes $n$ on GPQA-Diamond.}
\label{tab:mistralG-n-sweep-2}
\begin{center}
\resizebox{\linewidth}{!}{%
\begin{tabular}{c|ccccccccccc}
\hline
\bf $n$ & 2 & 3 & 4 & 5 & 6 & 7 & 8 & 9 & 10 & 11 & 12 \\
\hline
Mistral & \makecell{42.67\\$\pm$2.94} & \makecell{42.33\\$\pm$2.58} & \makecell{43.67\\$\pm$1.97} & \makecell{43.33\\$\pm$2.34} & \makecell{43.50\\$\pm$0.84} & \makecell{44.17\\$\pm$1.33} & \makecell{44.33\\$\pm$1.03} & \makecell{43.83\\$\pm$0.75} & \makecell{43.83\\$\pm$0.41} & \makecell{43.50\\$\pm$0.55} & \makecell{44.33\\$\pm$1.03} \\
\hline
\end{tabular}%
}
\vspace{4pt}

\resizebox{\linewidth}{!}{%
\begin{tabular}{c|cccccccccc|c}
\hline
\bf $n$ & 13 & 14 & 15 & 16 & 17 & 18 & 19 & 20 & 21 & 22 & \bf Cost \\
\hline
Mistral & \makecell{44.17\\$\pm$0.98} & \makecell{43.67\\$\pm$0.82} & \makecell{44.50\\$\pm$1.38} & \makecell{44.17\\$\pm$0.75} & \makecell{44.33\\$\pm$0.52} & \makecell{44.83\\$\pm$1.17} & \makecell{44.67\\$\pm$1.37} & \makecell{44.83\\$\pm$1.47} & \makecell{44.67\\$\pm$2.25} & \makecell{44.50\\$\pm$1.87} & 100.49\% \\
\hline
\end{tabular}%
}
\end{center}
\end{table}

\begin{table}[htbp]
\caption{Consistency Accuracy (\%) of Fuse-Extension across different sample sizes $n$ on GPQA-Diamond.}
\label{tab:fuseG-extension-n-sweep-2}
\begin{center}
\resizebox{\linewidth}{!}{%
\begin{tabular}{c|ccccccccccc|c}
\hline
\bf $n$ & 2 & 3 & 4 & 5 & 6 & 7 & 8 & 9 & 10 & 11 & 12 & \bf Cost \\
\hline
Fuse-Extension & \makecell{58.50\\$\pm$2.26} & \makecell{61.33\\$\pm$3.56} & \makecell{61.17\\$\pm$3.19} & \makecell{62.67\\$\pm$2.16} & \makecell{60.67\\$\pm$2.58} & \makecell{58.00\\$\pm$2.90} & \makecell{57.83\\$\pm$3.25} & \makecell{57.33\\$\pm$4.84} & \makecell{57.33\\$\pm$3.27} & \makecell{58.83\\$\pm$1.94} & \makecell{59.83\\$\pm$3.06} & 96.24\% \\
\hline
\end{tabular}%
}
\end{center}
\end{table}

\begin{table}[htbp]
\caption{Consistency Accuracy (\%) of Best-Extension across different sample sizes $n$ on GPQA-Diamond.}
\label{tab:bestG-extension-n-sweep-2}
\begin{center}
\resizebox{\linewidth}{!}{%
\begin{tabular}{c|cccccccccc|c}
\hline
\bf $n$ & 2 & 3 & 4 & 5 & 6 & 7 & 8 & 9 & 10 & 11 & \bf Cost \\
\hline
Best-Extension & \makecell{59.17\\$\pm$4.40} & \makecell{61.33\\$\pm$4.50} & \makecell{60.67\\$\pm$3.39} & \makecell{61.83\\$\pm$2.64} & \makecell{62.83\\$\pm$2.04} & \makecell{63.17\\$\pm$2.48} & \makecell{61.50\\$\pm$2.81} & \makecell{62.17\\$\pm$1.94} & \makecell{62.17\\$\pm$2.32} & \makecell{62.83\\$\pm$1.72} & 99.14\% \\
\hline
\end{tabular}%
}
\end{center}
\end{table}

\begin{table}[htbp]
\caption{Consistency Accuracy (\%) of Grok across different sample sizes $n$ on MMLU-Pro.}
\label{tab:grokPRO-n-sweep-3}
\begin{center}
\resizebox{\linewidth}{!}{%
\begin{tabular}{c|cccccccc|c}
\hline
\bf $n$ & 2 & 3 & 4 & 5 & 6 & 7 & 8 & 9 & \bf Cost \\
\hline
Grok & \makecell{57.83\\$\pm$1.47} & \makecell{59.00\\$\pm$1.41} & \makecell{60.00\\$\pm$0.63} & \makecell{60.17\\$\pm$1.47} & \makecell{61.00\\$\pm$1.55} & \makecell{61.00\\$\pm$1.41} & \makecell{60.50\\$\pm$1.52} & \makecell{60.33\\$\pm$1.37} & 115.57\% \\
\hline
\end{tabular}%
}
\end{center}
\end{table}

\begin{table}[htbp]
\caption{Consistency Accuracy (\%) of GPT across different sample sizes $n$ on MMLU-Pro.}
\label{tab:gptPRO-n-sweep-3}
\begin{center}
\resizebox{\linewidth}{!}{%
\begin{tabular}{c|ccccccccccccccc|c}
\hline
\bf $n$ & 2 & 3 & 4 & 5 & 6 & 7 & 8 & 9 & 10 & 11 & 12 & 13 & 14 & 15 & 16 & \bf Cost \\
\hline
GPT & \makecell{45.50\\$\pm$3.02} & \makecell{45.83\\$\pm$2.71} & \makecell{47.33\\$\pm$1.21} & \makecell{47.00\\$\pm$0.00} & \makecell{47.50\\$\pm$1.76} & \makecell{47.50\\$\pm$0.84} & \makecell{47.33\\$\pm$0.52} & \makecell{47.83\\$\pm$1.60} & \makecell{47.83\\$\pm$1.33} & \makecell{48.50\\$\pm$1.05} & \makecell{47.67\\$\pm$0.82} & \makecell{47.33\\$\pm$1.03} & \makecell{47.17\\$\pm$1.33} & \makecell{47.67\\$\pm$1.63} & \makecell{47.33\\$\pm$1.63} & 103.85\% \\
\hline
\end{tabular}%
}
\end{center}
\end{table}

\begin{table}[htbp]
\caption{Consistency Accuracy (\%) of Mistral across different sample sizes $n$ on MMLU-Pro.}
\label{tab:mistralPRO-n-sweep-3}
\begin{center}
\resizebox{\linewidth}{!}{%
\begin{tabular}{c|ccccccccc}
\hline
\bf $n$ & 2 & 3 & 4 & 5 & 6 & 7 & 8 & 9 & 10 \\
\hline
Mistral & \makecell{57.50\\$\pm$2.17} & \makecell{58.83\\$\pm$1.72} & \makecell{58.00\\$\pm$1.79} & \makecell{58.33\\$\pm$2.25} & \makecell{58.67\\$\pm$3.39} & \makecell{58.33\\$\pm$2.25} & \makecell{58.33\\$\pm$2.50} & \makecell{58.17\\$\pm$2.64} & \makecell{58.67\\$\pm$2.34} \\
\hline
\end{tabular}%
}
\vspace{4pt}

\resizebox{\linewidth}{!}{%
\begin{tabular}{c|ccccccccc|c}
\hline
\bf $n$ & 11 & 12 & 13 & 14 & 15 & 16 & 17 & 18 & 19 & \bf Cost \\
\hline
Mistral & \makecell{58.83\\$\pm$2.04} & \makecell{58.83\\$\pm$2.04} & \makecell{58.50\\$\pm$1.38} & \makecell{57.83\\$\pm$1.94} & \makecell{58.33\\$\pm$1.21} & \makecell{57.83\\$\pm$1.47} & \makecell{57.50\\$\pm$1.05} & \makecell{57.83\\$\pm$1.72} & \makecell{58.00\\$\pm$1.10} & 103.16\% \\
\hline
\end{tabular}%
}
\end{center}
\end{table}

\begin{table}[htbp]
\caption{Consistency Accuracy (\%) of Fuse-Extension across different sample sizes $n$ on MMLU-Pro.}
\label{tab:fusePRO-extension-n-sweep-3}
\begin{center}
\resizebox{\linewidth}{!}{%
\begin{tabular}{c|ccccccccccc|c}
\hline
\bf $n$ & 2 & 3 & 4 & 5 & 6 & 7 & 8 & 9 & 10 & 11 & 12 & \bf Cost \\
\hline
Fuse-Extension & \makecell{57.67\\$\pm$2.58} & \makecell{59.00\\$\pm$1.55} & \makecell{59.83\\$\pm$1.33} & \makecell{60.83\\$\pm$1.17} & \makecell{62.33\\$\pm$2.25} & \makecell{64.33\\$\pm$2.07} & \makecell{64.33\\$\pm$2.25} & \makecell{66.17\\$\pm$2.32} & \makecell{65.83\\$\pm$2.23} & \makecell{64.50\\$\pm$1.76} & \makecell{64.50\\$\pm$2.07} & 99.23\% \\
\hline
\end{tabular}%
}
\end{center}
\end{table}

\begin{table}[htbp]
\caption{Consistency Accuracy (\%) of Best-Extension across different sample sizes $n$ on MMLU-Pro.}
\label{tab:bestPRO-extension-n-sweep-3}
\begin{center}
\resizebox{\linewidth}{!}{%
\begin{tabular}{c|cccccccccc|c}
\hline
\bf $n$ & 2 & 3 & 4 & 5 & 6 & 7 & 8 & 9 & 10 & 11 & \bf Cost \\
\hline
Best-Extension & \makecell{57.67\\$\pm$2.07} & \makecell{59.17\\$\pm$1.83} & \makecell{57.83\\$\pm$2.04} & \makecell{59.00\\$\pm$2.45} & \makecell{60.50\\$\pm$2.35} & \makecell{62.50\\$\pm$3.08} & \makecell{63.33\\$\pm$3.20} & \makecell{64.17\\$\pm$3.06} & \makecell{63.67\\$\pm$2.50} & \makecell{63.17\\$\pm$3.31} & 99.88\% \\
\hline
\end{tabular}%
}
\end{center}
\end{table}

\end{document}